\documentclass[lettersize,journal]{IEEEtran}
\usepackage{amsmath,amsfonts}
\usepackage{algorithmic}
\usepackage{algorithm}
\usepackage{array}
\usepackage[caption=false,font=normalsize,labelfont=sf,textfont=sf]{subfig}
\usepackage{textcomp}
\usepackage{stfloats}
\usepackage{url}
\usepackage{verbatim}
\usepackage{graphicx}
\usepackage{cite}
\usepackage{multirow}
\usepackage{pifont}
\newtheorem{definition}{Definition}[section]
\newtheorem{hypothesis}{Hypothesis}[section]

\usepackage{booktabs}

\usepackage{color}

\begin{document}

\title{Correlation or Causation: Analyzing the Causal Structures of LLM and LRM Reasoning Process}

% \author{Zhizhang~Fu, Guangsheng~Bao, Hongbo~Zhang, Chenkai~Hu and Yue~Zhang,~\IEEEmembership{Member,~IEEE}%
\author{
  Zhizhang~Fu\textsuperscript{*}, 
  Guangsheng~Bao\textsuperscript{*}, 
  Hongbo~Zhang, 
  Chenkai~Hu, 
  Yue~Zhang\textsuperscript{\textdagger},~
  \IEEEmembership{Member,~IEEE}
}

% \thanks{Manuscript submitted to TASLP.}
% \thanks{Guangsheng Bao is with the School of Engineering, Westlake University, Hangzhou 310024, China (email: baoguangsheng@westlake.edu.cn).}% 
% \thanks{Yue Zhang is with the School of Engineering, Westlake University, Hangzhou 310024, China, and also with the Institute of Advanced Technology, Westlake Institute of Advanced Study, Hangzhou 310024, China (email: yue.zhang@wias.org.cn).}}


\markboth{IEEE/ACM TRANSACTIONS ON AUDIO, SPEECH, AND LANGUAGE PROCESSING, VOL. XX, NO. X, XXXXX 2025}
{Shell \MakeLowercase{\textit{et al.}}: A Sample Article Using IEEEtran.cls for IEEE Journals}

\maketitle

\footnotetext[1]{Contributed equally to this work as co-first authors}
\footnotetext[2]{Corresponding author: yue.zhang@wias.org.cn}

\begin{abstract}
LLMs suffer from critical reasoning issues such as unfaithfulness, bias, and inconsistency, since they lack robust causal underpinnings and may rely on superficial correlations rather than genuine understanding. Successive LRMs have emerged as a promising alternative, leveraging advanced training techniques such as reinforcement learning (RL) and distillation to improve task accuracy. However, the impact of these training methods on causality remains largely unexplored. In this study, we conduct a systematic causal analysis on LLMs and LRMs, examining structural causal models (SCMs) of four key variables: problem instruction (Z), thinking process (T), reasoning steps (X), and answer (Y). Our findings reveal that RLVR-trained LRMs exhibit enhanced causal reasoning capabilities, aligning more closely with ideal causal structures, while LLMs and distilled LRMs fail to address causality-related deficiencies. Our further investigation indicates that RLVR reduces spurious correlations and strengthens genuine causal patterns, thereby mitigating unfaithfulness and bias. In addition, our inspection on the dynamics of the RLVR training process observes a high correlation between reduced spurious features and improved causal structures, where the causal relationships consistently improve in the training process. This study contributes to the understanding of causality in reasoning models, highlights the critical role of RLVR in enhancing causal reasoning, and provides insights for designing future AI systems with stronger causal foundations. We release our code and data at https://github.com/Harryking1999/CoT\_Causal\_Analysis.
\end{abstract}


\section{Introduction}
\IEEEPARstart{L}{arge} Language Models (LLMs; e.g., GPT4~\cite{achiam2023gpt}, DeepSeek-V3~\cite{deepseekai2024deepseekv3technicalreport}) and successive Large Reasoning Models (LRMs; e.g., o1~\cite{openai2025o1preview}, DeepSeek-R1~\cite{guo2025deepseek}) have demonstrated impressive reasoning capabilities, producing explainable reasoning processes (chain of thoughts) \cite{wei2022chain,kojima2022large,wang2023plan}. LRMs differ from LLMs in further giving a thinking process before explicit reasoning. Despite showing effectiveness in math and other tasks, CoT reasoning processes often suffer from critical issues such as unfaithfulness, bias, and inconsistency \cite{lanham2023measuring,turpin2023language,bentham2024chain,jin2024impact,bao2025likely}, which has caused concerns about its reliability~\cite{paul2024making,pfau2024let,barez2025chain}. Empirical evidence shows that correct CoTs may lead to incorrect answers, and incorrect CoTs to correct answers \cite{bao2025likely}.

\begin{figure}[t]
    \centering\small
    \includegraphics[width=0.8\linewidth]{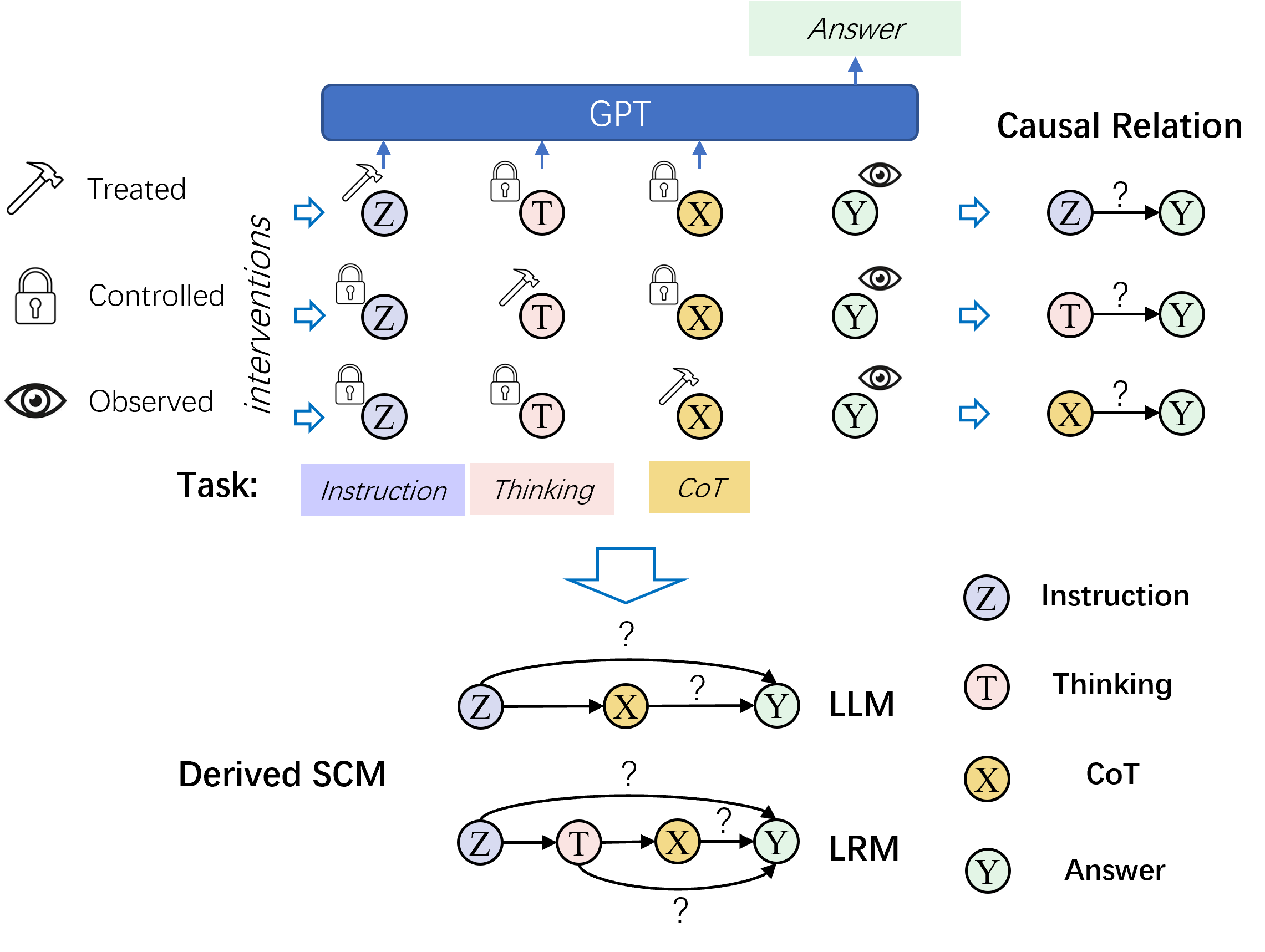}
    \caption{Causal Analysis to derive underlying causal structure of LLMs and LRMs. We conduct treated experiments that interfere (treat) one variable while freeze (control) others, so that we can observe their effects.}
    \label{fig:head}
\end{figure}

% The necessity of causal analysis
Intuitively, robust reasoning is rooted in a causal chain of thoughts, not just superficial correlations between thoughts. An ideal reasoning process often involves identifying causal relationships to make sense of existing facts and to predict future occurrences~\cite{pearl2009causality}, which leads to stronger robustness and generalization~\cite{zhou2024towards}. Therefore, analyzing LLM and LRM reasoning from a causal perspective is essential for deeper insights into their behavior and for advancing the reasoning models, answering essential questions including \emph{when} and \emph{how} a causal reasoning process happens, which existing work does not consider ~\cite{jin2024impact,harsha2024hardness,yee2024dissociation}.

\begin{figure*}[t]
    \centering\small
    \includegraphics[width=0.8\linewidth]{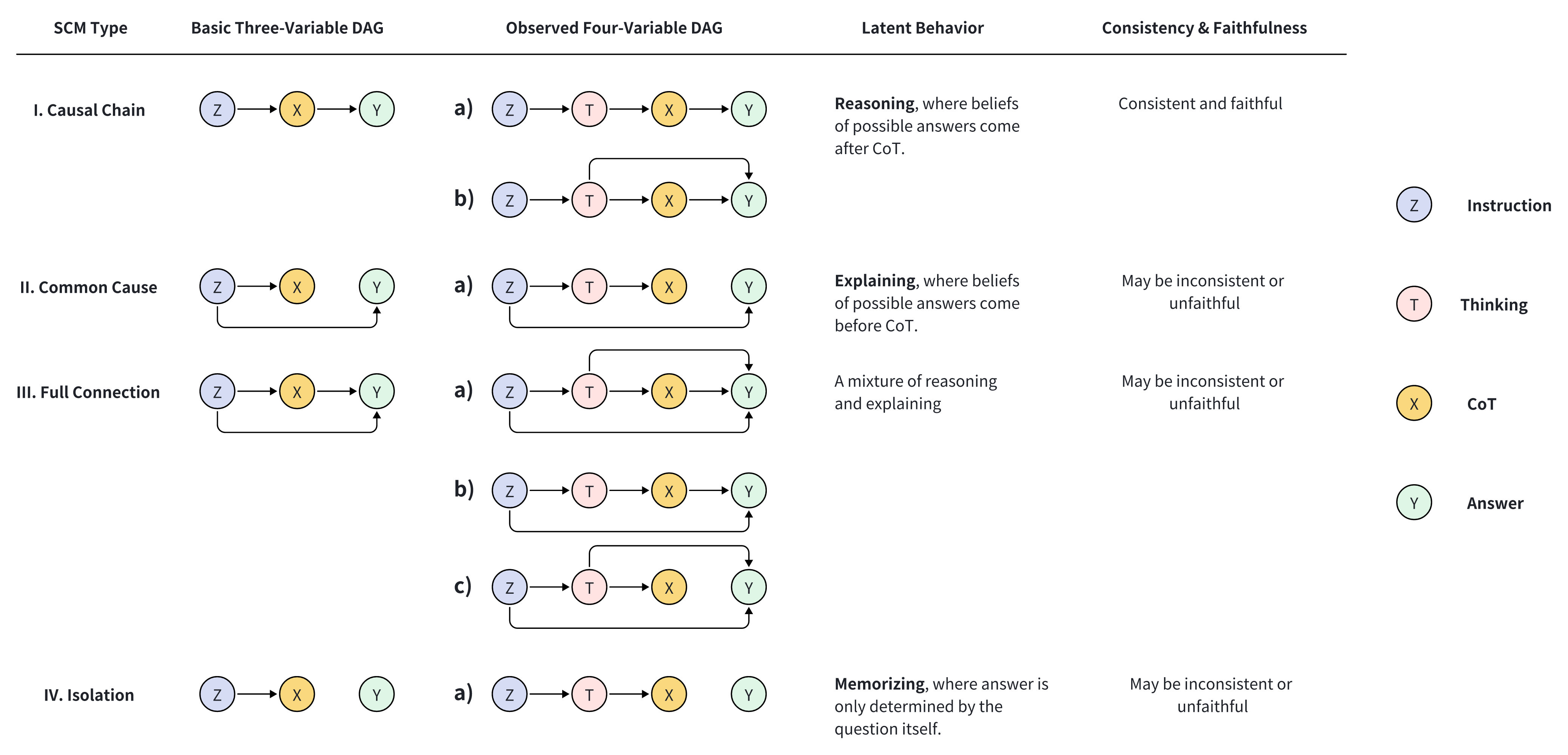}
    \caption{Structural causal model (SCM) types and their latent behavior, consistency, and faithfulness. We use the three-variable DAGs for LLMs while the four-variable DAGs for LRMs.}
    \label{fig:SCM}
\end{figure*}

% Our causal analysis
We conduct a causal analysis for CoT, dividing the reasoning process into four parts, each represented by a random variable. Specifically, as Figure \ref{fig:head} illustrates, they are problem instruction ($Z$), thinking process (implicit CoT, $T$, for LRM only), reasoning steps (explicit CoT, $X$), and final conclusion (answer, $Y$). We infer the causal relationships between these variables and the answer $Y$, where a structural causal model (SCM) can be derived for the reasoning process. Typically, we identify four major types and seven minor types of causal structures, as Figure \ref{fig:SCM} shows. An ideal reasoning process is represented as a causal chain (type-I), where the instruction determines the reasoning, and the reasoning decides the answer. In this case, the model is faithfully reasoning, which produces consistent responses. In contrast, when a reasoning process has a common cause (type-II) structure, it is actually explaining a latent belief of the answer, which may produce unfaithful and inconsistent responses.

% Experiments and findings
We evaluate various LLMs and LRMs on a range of reasoning tasks (Section \ref{sec:lrm_scms}). Experimental results show that CoT of LLMs generally does not possess an ideal causal structure. In contrast, LRMs generally achieve higher task accuracy compared to LLMs, but their causal structures vary significantly depending on the training method. Distilled LRMs~\cite{guo2025deepseek}, which rely on supervised fine-tuning (SFT)~\cite{ouyang2022training,wei2022finetuned}, do not exhibit improved causal structures and often share similar causal deficiencies as LLMs. LRMs trained with reinforcement learning (i.e., reinforcement learning with verifiable rewards, RLVR)~\cite{shaoDeepSeekMathPushingLimits2024, zhengGroupSequencePolicy2025, yuDAPOOpenSourceLLM2025, liuUnderstandingR1ZeroLikeTraining2025, chuGPGSimpleStrong2025} demonstrate enhanced causality, aligning more closely with the ideal causal chain structure. These findings suggest that different LLM training paradigms may influence the causal reasoning capabilities differently.

% Analysis on training techniques
We analyze various techniques that could potentially influence the causal structures, including in-context learning (ICL), instruction-tuning, reinforcement learning with human feedback (RLHF), long CoT distillation, and RLVR (Section~\ref{sec:rl_extent}). We find that RLVR substantially enhances causal structures, whereas other techniques have a minimal impact. Specifically, ICL could strengthen the causal relationships with limited effects, while instruction-tuning, RLHF, and distillation weaken the causal structures. Among all the techniques, RLVR is the strongest for achieving ideal reasoning structures.

% Dynamics of RL Training and Causality  
To further understand the impact of RLVR on causality, we investigate the dynamics of causal structure evolution during RLVR training (Section~\ref{sec:rlvr_process}). Our experiments show that RLVR consistently improves the causal relationships over the course of training, by reducing spurious correlations and strengthening models' genuine causal patterns, leading to more robust reasoning models. This dynamic improvement highlights the potential of RLVR as a powerful tool for enhancing causality in reasoning models, addressing key issues such as unfaithfulness, bias, and inconsistency.

% Summary of Contributions
This manuscript significantly extends our conference paper~\cite{bao2025likely} which focuses on the causality of LLM CoT. First, we extend causal analysis from LLMs to LRMs, revealing that LRMs exhibit enhanced causal reasoning capabilities. Second, we systematically investigate various learning paradigms, finding that RLVR substantially enhances causal structures. Finally, we discuss why RLVR improves causality by conducting extensive empirical experiments. These investigations and findings offer valuable insights for the development of more reliable and causally robust reasoning models. Our findings pave the way for future research on causality-driven LLM systems and highlight the importance of RL in advancing reasoning capabilities.

\section{Related Work}
% Recent work:
% The Illusion of Thinking: Understanding the Strengths and Limitations of Reasoning Models via the Lens of Problem Complexity.
% Comment on The Illusion of Thinking: Understanding the Strengths and Limitations of Reasoning Models via the Lens of Problem Complexity
% Does reinforcement learning really incentivize reasoning capacity in LLMs beyond the baseline?

% include LLM work and LLM performance analysis
% first, LRM has achieved remarkable results.

%- list famous/SOTA models and their performance on reasoning related tasks.
%  models: Anthropic’s Claude 4, DeepSeek-R1, Google’s Gemini 2.5 Pro, xAI’s Grok 4, and OpenAI’s GPT-5 and o-series (o1, o3, o4)
%  benchmarks: MMLU-Pro, GPQA Diamond, Humanity's Last Exam, LiveCodeBench, SciCode, AIME, IFBench, AA-LCR
Various reasoning techniques have been proposed to enhance the reasoning ability of LLMs \cite{chu2023survey,reasoning2024survey}. Chain-of-thought (CoT) prompting \cite{wei2022chain}, as an early study elicits reasoning in LLMs, inspires numerous further investigations. Specifically, self-consistency \cite{wang2022self} votes the major decision from multiple reasoning paths, Tree-of-thought \cite{yao2023tree} searches the most confident reasoning path in a tree, and Graph-of-thought \cite{yao2023beyond} represents the thoughts as graph nodes and combines thoughts non-sequentially. Advanced CoT methods, like Faithful CoT \cite{lyu2023faithful} and Constraint CoT \cite{vacareanu2024general}, are further proposed to improve reasoning capabilities. In this paper, we focus on the very basic chain of thought to understand the underlying mechanism of how LLMs do reasoning, leaving the analysis of advanced methods for the future.
% Faithful CoT \citep{lyu2023faithful} first translates the question to a symbolic reasoning problem and then solves it using a deterministic solver. \citet{vacareanu2024general} introduces constraints to the reasoning steps to improve the reasoning capabilities. 

Recently, researchers have introduced Large Reasoning Models (LRMs) specifically designed for reasoning, achieving significant progress on reasoning problems. Leading models, including Claude 4~\cite{claude42025introduce}, DeepSeek-R1~\cite{guo2025deepseek}, Gemini 2.5 Pro~\cite{comaniciGemini25Pushing2025}, Qwen 3~\cite{qwen3}, Grok 4~\cite{xai2025grok4}, OpenAI o3~\cite{openai2025o3} and GPT-5~\cite{openai2025gpt5}, now set the pace on reasoning benchmarks like MMLU-PRO~\cite{wangMMLUProMoreRobust2024}, GPQA~\cite{reinGPQAGraduateLevelGoogleProof2023} and AIME~\cite{patelAIMEAISystem2024}. Collectively, these results indicate rapid progress toward more reliable general reasoning competence in diverse tasks. However, the reason for LRMs achieving stronger results in reasoning has not been understood from a causality perspective. Our work fills this gap.
% For example, GPT-5 achieved 81.7\% accuracy in MMLU-Pro \cite{wangMMLUProMoreRobust2024}, , Grok reached 87.5\% in GPQA Diamond \cite{reinGPQAGraduateLevelGoogleProof2023}, and GPT-5 achieves a 93.4\% in AIME \cite{patelAIMEAISystem2024}. 

%second, discussion of distillation and RL based LRMs
%- different post-training techniques to enhance reasoning capbilities
    % - mention SFT and this work (?SFT-S1) - https://arxiv.org/abs/2501.19393
    % - focus on distillation
    % - list pros/cons
%- list the pros/cons of RL (o1-like replication) - focus on RLVR（GRPO, GxPO）, need to mention MCTS(list some work, no detail, deepseek-r1 paper mention this not work)
% - differentiate between distill/RL

Among the LRM methods, OpenAI initially leverages large-scale reinforcement learning~\cite{openai2025o1preview}, marking the beginning of the LRM era. Following this breakthrough, the community has conducted extensive researches to reproduce o1-like LRMs, typically employing various RL algorithms for training and Monte Carlo Tree Search for test-time scaling~\cite{qin2024o1replicationjourneystrategic, zhang2024o1codero1replicationcoding, wang2024openropensourceframework}. Since the release of DeepSeek-R1 \cite{guo2025deepseek}, Reinforcement Learning with Verifiable Rewards (RLVR) has garnered considerable attention, particularly through the Group Relative Policy Optimization (GRPO) algorithm \cite{shaoDeepSeekMathPushingLimits2024}. The efficacy of GRPO in enhancing reasoning capabilities has spawned several variants, including Group Sequence Policy Optimization (GSPO) implemented in Qwen models \cite{zhengGroupSequencePolicy2025}, DAPO \cite{yuDAPOOpenSourceLLM2025}, Dr.GRPO \cite{liuUnderstandingR1ZeroLikeTraining2025}, and GPG \cite{chuGPGSimpleStrong2025}. RLVR has become a prominent post-training methodology for training LRMs.
% Reinforcement learning has emerged as a prominent post-training methodology for enhancing large language model capabilities \cite{kumarLLMPostTrainingDeep2025}. OpenAI demonstrated substantial performance improvements through post-training with chain-of-thought reasoning and reinforcement learning techniques \cite{openai2025o1preview}. The research community has adopted various classical RL algorithms for training large reasoning models (LRMs), including Direct Preference Optimization (DPO) \cite{xieMonteCarloTree2024}, Proximal Policy Optimization (PPO) \cite{schulmanProximalPolicyOptimization2017}, and Monte Carlo Tree Search (MCTS) \cite{gaoInterpretableContrastiveMonte2024}. Following the release of DeepSeek-R1 \cite{guo2025deepseek}, Reinforcement Learning with Verifiable Rewards (RLVR) has garnered considerable attention, particularly through the Group Relative Policy Optimization (GRPO) algorithm \cite{shaoDeepSeekMathPushingLimits2024} and various reward modeling techniques \cite{chenPasskTrainingAdaptively2025}. The efficacy of GRPO in enhancing reasoning capabilities has spawned several variants, including Group Sequence Policy Optimization (GSPO) implemented in Qwen models \cite{zhengGroupSequencePolicy2025}, DAPO \cite{yuDAPOOpenSourceLLM2025}, Dr.GRPO \cite{liuUnderstandingR1ZeroLikeTraining2025}, and GPG \cite{chuGPGSimpleStrong2025}. These developments indicate that RLVR has become the predominant methodology for training large reasoning models.

Meanwhile, supervised fine-tuning (SFT) represents another prevalent training paradigm for LRMs. Muennighoff et al.~\cite{muennighoffS1SimpleTesttime2025} demonstrate that SFT with a small amount of high-quality data can achieve superior performance compared to o1-preview~\cite{openai2025o1preview} on mathematical benchmarks. Concurrently, distillation has emerged as a mainstream SFT approach in recent LRM training. This method leverages the outputs of larger-scale models to empower smaller models with reasoning capabilities~\cite{guo2025deepseek}, enabling dramatic improvements in reasoning performance that approach the levels of their larger counterparts.

Despite the remarkable achievements of LRMs, some studies have questioned whether they engage in genuine reasoning. For example, Shojaee et al.~\cite{shojaee2025illusion} demonstrate that LRMs exhibit advantages in medium-complexity tasks, underperform compared to standard LLMs in low-complexity tasks, and both model types experience complete collapse when faced with high-complexity tasks. Additionally, there is considerable debate regarding the relative merits of the two primary training paradigms for LRMs: RL and distillation. On the one hand, RL demonstrates unparalleled advantages over other paradigms in training large-scale LRMs; on the other hand, for smaller and less capable models, only distillation can introduce new knowledge and rapidly improve performance, whereas RL can merely enable models to better utilize their existing knowledge, being constrained by the original model's knowledge limitations~\cite{yue2025doesreinforcementlearningreally}. In this work, we analyze the relative merits of distillation and RL from a causal perspective.

Causality represents the fundamental building blocks of both physical reality and human understanding, transcending mere statistical associations to capture the underlying mechanisms that govern phenomena. Moreover, causal relationships exhibit greater robustness and stability than probabilistic relationships, remaining ``invariant to changes in all mechanisms" and providing the foundation for genuine understanding rather than mere pattern recognition~\cite{pearl2009causality}. This highlights the importance of performing causal assessment on current LLMs. However, existing studies on the causal reasoning capabilities of LLM mainly focus on variables described in natural language, like benchmark ATOMIC \cite{sap2019atomic}, CLadder \cite{jin2023cladder}, and Corr2Cause \cite{jin2023can}. ATOMIC focuses on the if-then relations of variables like ``{if X pays Y, Y will probably return}''. CLadder requires the identification of variables and their causal relations from the language context prior to inference. Corr2Cause determines the causal structure between variables according to a group of correlational statements.
On multiple causal benchmarks, \cite{kiciman2023causal} finds that LLMs achieve good accuracy and hypothesizes that LLMs can use their collected knowledge to generate causal graphs from natural language, while \cite{zevcevic2023causal} conjectures that a successful causal inference relies on a pre-learned meta-SCM that stores the related causal facts in natural language. 
Unlike these studies, where variables represent targets in the question domain, we investigate the causality between four known variables, instruction, CoT, thinking and answer, which do not represent any question-specific target, but only abstractive components of the chain-of-thought reasoning.

\section{Causal Analysis}
As shown in Figure~\ref{fig:head}, we take an intervention-based approach to conduct causal analysis for CoT, where a causal graph is constructed (Section~\ref{random_variables}), and causal relations are calculated (Section~\ref{identification}), so that the CoT patterns can be categorized into causal structure prototypes (Section~\ref{scm_type}), and different prototypes exhibit distinct CoT behaviors (Section~\ref{scm_behavior}).
\subsection{Random Variables}
\label{random_variables}
We model the reasoning process of LLMs in three random variables and LRMs in four random variables, involving instruction ($Z$), CoT ($X$) and answer ($Y$) for LLMs and an additional thinking ($T$) for LRMs, described as follows.

\textbf{Instruction ($Z$)} generally consists of a task description, a question formulation, several examples, to guide the models toward an appropriate response. Because few-shot prompting is not recommended in some reasoning models~\cite{guo2025deepseek}, our instructions follow the simplest style, directly describing the question with a zero-shot setting. 

\textbf{Thinking ($T$; for LRM only)} is a new component introduced in LRMs. It consists of a long chain-of-thought (CoT), characterized by extensive exploration, feasible reflection, and deep reasoning~\cite{chen2025towards}. This stage provides an internal process by which the model thoroughly analyzes the problem before generating the explicit CoT or answer. In this work, we investigate whether the inclusion of the thinking stage has any impact on the overall causality of reasoning in the models.

\textbf{CoT ($X$)} is the step-by-step reasoning process of LLMs, which is often questioned for lacking faithfulness. We therefore set CoT as an independent variable and verify its causal relationship with Answer.

\textbf{Answer ($Y$)} represents the final step of the reasoning process, which answers the question. Unlike the two variables above, which work as either independent variables or constants, answer consistently serves as a dependent variable in experiments. All experiments focus on studying the relationships between other variables and the answer, which enables us to evaluate the reasoning causality structures.

\subsection{Identification of Causal Relations}
\label{identification}
In autoregressive language models, a token can depend on any tokens on the left, therefore presenting arbitrary subgraphs of fully connected structural causal models (SCMs). To figure out the specific SCM a model follows, we employ treatment experiments~\cite{rubin1974estimating,angrist1995identification}, with each time intervening only one independent variable and observing answer ($Y$) as the dependent variable. Below are several key definitions: 

\begin{definition}[\textbf{Cause-Effect Interventions}]
    Suppose that the SCM $\mathcal{G}$ entails a distribution $P_{X,Y}$ with $N_X,N_Y \overset{iid}{\sim} \mathcal{N}(0,1)$. Then we intervene on $X$ to change the distribution of $Y$
    \begin{equation}
    \small
    \begin{split}
        P_Y^{do(X)} = P(Y|do(X)).
    \label{eq:scm-inter}
    \end{split}
    \end{equation}
\end{definition}

By intervening on an independent variable and observing the dependent variable before and after the intervention, we can determine whether they are causally related.

\begin{definition}[\textbf{Average Treatment Effect; ATE}]
    An ATE \cite{rubin1974estimating} represents the effect of an intervention, which compares the distributions of the target variable $Y$ with and without a treatment.
    \begin{equation}
    \small
    \begin{split}
        \textit{ATE} = E(Y|do(X)) - E(Y).
    \label{eq:ate}
    \end{split}
    \end{equation}
\end{definition}

We use the significance of ATE to determine the causal relationship between two variables. Specifically, we employ McNemar's test~\cite{mcnemar1947note} for assessing the significance. Since our interventions are all negative, we take the absolute value of negative ATEs and set positive ATEs to zero, in order to avoid confounding the causal relationship.

\begin{definition}[\textbf{Relative Average Treatment Effect; R-ATE}]
    We introduce R-ATE to measure the proportion of the actual treatment effect relative to the maximum possible treatment effect. It indicates what percentage of the total achievable effect is actually realized.
    \begin{equation}
    \small
    \begin{split}
        \textit{R-ATE} = \frac{\text{ATE}}{\text{Possible Max ATE}} \\
    \label{eq:rate}
    \end{split}
    \end{equation}
\end{definition}

The value of ATE is not only influenced by underlying causal mechanisms we aim to measure, but also by other irrelevant factors. For instance, if model A changes from 0.9 to 0, and model B changes from 0.7 to 0 under the same intervention, both are completely intervened, but their ATEs differ, making ATE incomparable across different models. R-ATE addresses this issue by enabling fair comparisons across models. However, R-ATE is still not comparable across different intervention methods.

Particularly in our experiments, we test different hypotheses depending on the SCM structure. For three-variable SCMs, we test two hypotheses: whether instruction causes the answer, and whether CoT causes the answer. For four-variable SCMs, we test \textbf{three hypotheses}: \emph{whether instruction, thinking and CoT, causes the answer}.

\begin{hypothesis}[\textbf{Instruction causes Answer}]
    Given a constant CoT $X$ and Thinking $T$
    \begin{equation}
    \small
    \left\{
    \begin{array}{ll}
      H_0: \text{ATE} = 0, & \text{Instruction does not cause Answer,} \\
      H_1: \text{ATE} \neq 0, & \text{Instruction causes Answer,}
    \end{array}
    \right.
    \end{equation}
    where $\text{ATE} = E(Y|T,X,do(Z)) - E(Y|T,X)$.
\label{hyp:instruction_cause_answer}
\end{hypothesis}

\begin{hypothesis}[\textbf{CoT causes Answer}]
    Given a constant Instruction $Z$ and Thinking $T$
    \begin{equation}
    \small
    \left\{
    \begin{array}{ll}
      H_0: \text{ATE} = 0, & \text{CoT does not cause Answer,} \\
      H_1: \text{ATE} \neq 0, & \text{CoT causes Answer,}
    \end{array}
    \right.
    \end{equation}
    where $\text{ATE} = E(Y|Z,T,do(X)) - E(Y|Z,T)$.
\label{hyp:cot_cause_answer}
\end{hypothesis}

\begin{hypothesis}[\textbf{Thinking causes Answer}]
    Given a constant Instruction $Z$ and CoT $X$
    \begin{equation}
    \small
    \left\{
    \begin{array}{ll}
      H_0: \text{ATE} = 0, & \text{Thinking does not cause Answer,} \\
      H_1: \text{ATE} \neq 0, & \text{Thinking causes Answer,}
    \end{array}
    \right.
    \end{equation}
    where $\text{ATE} = E(Y|Z,X,do(T)) - E(Y|Z,X)$.
\label{hyp:thinking_cause_answer}
\end{hypothesis}

To validate these hypotheses, we employ distinct intervention strategies for each. For Hypothesis~\ref{hyp:cot_cause_answer} and Hypothesis~\ref{hyp:thinking_cause_answer}, we intervene on the CoT by replacing it with randomized reasoning chains. For Hypothesis~\ref{hyp:instruction_cause_answer}, we intervene on the instruction with biases like ``{\it I think the answer is $w$}'', where $w$ represents an incorrect answer. The statistical significance of these hypotheses reveals the underlying SCM of each LLM in each task, and the R-ATE provides more fine-grained information as a continuous value rather than a binary classification like significance.

%===================================
\begin{table*}[h]
\centering
\begin{tabular}{ll|l|cccc|ccc}
\toprule
\multirow{2}{*}{\textbf{Model}} & \multirow{2}{*}{\textbf{Type}} & \multirow{2}{*}{\textbf{Setting}} & \multicolumn{4}{c|}{\textbf{Math}} & \multicolumn{3}{c}{\textbf{Logic}} \\
\cline{4-10}
 &  &  & \textbf{Add.} & \textbf{Mult.} & \textbf{Math500} & \textbf{GSM8K} & \textbf{ProofWriter} & \textbf{FOLIO} & \textbf{LogiQA} \\
\midrule
\multirow{6}{*}{GPT-4} & \multirow{6}{*}{LLM} & default & 0.998 & 0.816 & - & 0.946 & 0.708 & 0.686 & 0.688 \\
\cline{3-10}
 & & random CoT & 0.994 & 0.040 & - & 0.932 & 0.626 & 0.637 & 0.683 \\
 & & R-ATE: X$\rightarrow$Y & $0.60\%_{F}$ & $95.3\%_{T}$ & - & $1.7\%_{F}$ & $11.6\%_{T}$ & $7.1\%_{F}$ & $0.3\%_{F}$ \\
\cline{3-10}
 & & Instruction bias & 0.826 & 0.848 & - & 0.946 & 0.703 & 0.671 & 0.675 \\
 & & R-ATE: Z$\rightarrow$Y & $17.4\%_{T}$ & $1.2\%_{F}$ & - & $0.2\%_{F}$ & $0.7\%_{F}$ & $2.2\%_{F}$ & $1.5\%_{F}$ \\
\cline{3-10}
 &  & SCM Type & II & I & - & IV & I & IV & IV \\
\midrule
\multirow{9}{*}{Deepseek-R1} & \multirow{9}{*}{LRM} & default & 0.996 & 0.99 & 0.962 & - & 0.893 & 0.799 & 0.807 \\
\cline{3-10}
 &  & random Thinking & 0.994 & 0.988 & 0.956 & - & 0.887 & 0.784 & 0.806 \\
 &  & R-ATE: T$\rightarrow$Y & $0.2\%_{F}$ & $0.2\%_{F}$ & $0.6\%_{F}$ & - & $0.6\%_{F}$ & $1.9\%_{F}$ & $0.1\%_{F}$ \\
 \cline{3-10}
 &  & random CoT & 0.278 & 0.652 & 0.858 & - & 0.793 & 0.637 & 0.740 \\
 &  & R-ATE: X$\rightarrow$Y & $72.1\%_{T}$ & $34.1\%_{T}$ & $10.8\%_{T}$ & - & $11.2\%_{T}$ & $20.3\%_{T}$ & $8.3\%_{T}$ \\
\cline{3-10}
 &  & Instruction bias & 0.978 & 0.982 & 0.960 & - & 0.887 & 0.794 & 0.812 \\
 &  & R-ATE: Z$\rightarrow$Y & $1.8\%_{T}$ & $0.8\%_{F}$ & $0.2\%_{F}$ & - & $0.7\%_{F}$ & $0.6\%_{F}$ & $0_{F}$ \\
\cline{3-10}
 &  & SCM Type & III & I & I & - & I & I & I \\
\bottomrule
% \label{tab:SCM_identification_example}
\end{tabular}
\caption{Identification of causal structures in tasks running on GPT-4 and DeepSeek-R1, as representatives of LLMs and LRMs, respectively. We present task accuracy and R-ATE, where the subscript `{\it T/F}' denotes statistical significance with $p<0.01$ in McNemar's test. ``Default'' refers to the default CoT generated by the LLM; ``random CoT'' and ``random thinking'' indicate that we intervened with random content in parts of the CoT or thinking processes; ``instruction bias'' represents the introduction of bias into the instruction. Finally, we determine SCM types based on statistical significance.}
\label{tab:main_table}
\end{table*}

\subsection{SCM Types}
\label{scm_type}
% As shown in Figure~\ref{fig:SCM}, we illustrate the four basic SCM types using three-variable directed acyclic graphs (DAGs). To maintain consistency and comparability, we align the four-variable SCMs with the three-variable SCMs by omitting examination of the effect of $T$ on $Y$.
As shown in Figure~\ref{fig:SCM}, we define the four basic SCM types using three-variable directed acyclic graphs (DAGs). To maintain consistency and comparability, we align the four-variable SCMs with the three-variable SCMs by treating the combined effects of $T$ and $X$ on $Y$ as equivalent to the effect of $X$ on $Y$ in the three-variable case (where significance of either $T$ or $X$ on $Y$ corresponds to significance of $X$ on $Y$), while the effect of $Z$ on $Y$ remains identical across both frameworks. Therefore, the four-variable SCMs and three-variable SCMs capture the same underlying causal relationships, with the four-variable SCMs providing a finer-grained representation.

\textbf{Type I. Causal Chain} is the ideal SCM in our experimental setting, where the answer is caused by CoT and not by Instruction. This represents the desired behavior in which the model derives answers based on variables that should have a causal effect.

\textbf{Type II. Common Cause} signifies an SCM that Instruction causes the answer while CoT does not. In this case, CoT is more like an explanation to the models' hidden belief, instead of reasoning. 

\textbf{Type III. Full Connection} is the closest to the nature of autoregressive statistical models, where $Y$ depends on the $Z$ and $X$ in a left-to-right manner, capturing the causal mask property but representing statistical correlation rather than genuine causal relationships.

\textbf{Type IV. Isolation} represents that the answer is influenced solely by the question itself, where neither $Z$ nor $X$ determines the answer. Under this type of SCM, the model resembles memorizing the answer rather than genuinely performing reasoning.

\subsection{Connection between SCM, Latent Behavior, Consistency, and Faithfulness}
\label{scm_behavior}
Causal chain (type I) presents the ideal causal structure, where instruction ($\rightarrow$ thinking) $\rightarrow$ CoT $\rightarrow$ answer. Within this framework, Z is precluded from affecting Y, while both the presence and absence of T's effect on Y are reasonable. When a model works under causal chain, it does real reasoning, producing faithful and consistent output. In contrast, common cause (type II) represents an undesirable underlying structure, where the CoT is not a faithful reflection of models' answer-derivation process but a post-hoc explanation conditioned on the latent belief; on the other hand, the answer and the CoT are not necessarily consistent.
Generally speaking, all language models are fully connected (type III) to some extent due to the auto-regressive design. Thus, the real latent behavior of a LLM is most likely a mixture of reasoning and explaining. However, we can differentiate them using statistical significance as an indicator, obtaining the most likely underlying causal structures of an LLM on different tasks.
Additionally, isolation (type IV) represents a prototypical erroneous SCM, wherein the model's response is determined exclusively by the input question, remaining uninfluenced by any other variables within the causal framework. Under this structure, the model relies on memorized question-answer mappings rather than engaging in genuine reasoning processes.

\begin{table*}[h]
\centering
\begin{tabular}{l|cc|cccc|c}
\toprule
\textbf{Type} & \textbf{Model} & \textbf{Method} & \textbf{SCM-I (the ideal)} & \textbf{SCM-II} & \textbf{SCM-III} & \textbf{SCM-IV} & \textbf{SCM(I) \%} \\
\midrule
 \multirow{17}{*}{LLM} & \multirow{3}{*}{ChatGPT} & \multirow{3}{*}{RLHF} & \multirow{3}{*}{\shortstack{GSM8K \\ FOLIO}} & \multirow{3}{*}{\shortstack{Add.}} & \multirow{3}{*}{\shortstack{Mult. \\ ProofW. \\ LOGIQA}} &  \multirow{3}{*}{} &  \multirow{3}{*}{33\%}  \\ 
 & & & & & & \\  & & & & & & \\
& \multirow{3}{*}{GPT-4} & \multirow{3}{*}{RLHF} & \multirow{3}{*}{\shortstack{Mult. \\ ProofW.}} & \multirow{3}{*}{\shortstack{Add.}} & \multirow{3}{*}{} & \multirow{3}{*}{\shortstack{GSM8K \\ FOLIO \\ LOGIQA}} & \multirow{3}{*}{33\%}  \\  & & & & & & \\  & & & & & & \\
& \multirow{4}{*}{Llama-70B-Chat} & \multirow{4}{*}{RLHF} & \multirow{4}{*}{\shortstack{Add. \\ ProofW.}} & \multirow{4}{*}{} & \multirow{4}{*}{} & \multirow{4}{*}{\shortstack{Mult. \\ GSM8K \\ FOLIO \\ LOGIQA}} & \multirow{4}{*}{33\%}  \\  & & & & & & \\  & & & & & & \\ & & & & & & \\
& \multirow{4}{*}{Llama-7B-Chat} & \multirow{4}{*}{RLHF} & \multirow{4}{*}{} & \multirow{4}{*}{\shortstack{Add. \\ ProofW. \\ FOLIO \\ LOGIQA}} & \multirow{4}{*}{\shortstack{GSM8K}} & \multirow{4}{*}{\shortstack{Mult.}} & \multirow{4}{*}{0}  \\  & & & & & & \\  & & & & & & \\ & & & & & & \\
& \multirow{3}{*}{Qwen2.5-32B-Instruct} & \multirow{3}{*}{Instr.} & \multirow{3}{*}{\shortstack{Mult. \\ ProofW. \\ FOLIO}} & \multirow{3}{*}{} & \multirow{3}{*}{\shortstack{Add. \\ MATH500. \\ LOGIQA}} & \multirow{3}{*}{} & \multirow{3}{*}{50\%}  \\  & & & & & & \\  & & & & & & \\
\midrule
& & Avg. Ratio & 30\% & 20\% & 23\% & 27\% & 30\% \\
\midrule
\multirow{20}{*}{LRM} & \multirow{3}{*}{R1-Distill-Qwen-7B} & \multirow{3}{*}{Distill} & \multirow{3}{*}{\shortstack{Add.(a) \\ Mult.(b) \\ FOLIO(b)}} & \multirow{3}{*}{} & \multirow{3}{*}{\shortstack{MATH500(a) \\ ProofW.(a) \\ LOGIQA(c)}} & \multirow{3}{*}{} & \multirow{3}{*}{50\%}  \\  & & & & & & \\  & & & & & & \\
& \multirow{3}{*}{R1-Distill-Qwen-32B} & \multirow{3}{*}{Distill} & \multirow{3}{*}{\shortstack{Add.(a) \\ MATH500(b) \\ FOLIO(b)}} & \multirow{3}{*}{} & \multirow{3}{*}{\shortstack{Mult.(a) \\ ProofW.(a) \\ LOGIQA(a)}} & \multirow{3}{*}{} & \multirow{3}{*}{50\%}  \\  & & & & & & \\  & & & & & & \\
& \multirow{3}{*}{R1-Distill-Llama-8B} & \multirow{3}{*}{Distill} & \multirow{3}{*}{\shortstack{Mult.(a) \\ MATH500(b) \\ FOLIO(a)}} & \multirow{3}{*}{LOGIQA(a)} & \multirow{3}{*}{\shortstack{Add.(a) \\ ProofW.(b)}} & \multirow{3}{*}{} & \multirow{3}{*}{50\%}  \\  & & & & & & \\  & & & & & & \\
& \multirow{5}{*}{Deepseek-R1} & \multirow{5}{*}{misc.} & \multirow{5}{*}{\shortstack{Mult.(a) \\ MATH500(a) \\ ProofW.(a) \\ FOLIO(a) \\ LOGIQA(a)}} & \multirow{5}{*}{} & \multirow{5}{*}{Add.(b)} & \multirow{5}{*}{} & \multirow{5}{*}{\textbf{83\%}}  \\  & & & & & & \\  & & & & & & \\ & & & & & & \\ & & & & & & \\
& \multirow{6}{*}{QwQ-32B} & \multirow{6}{*}{RLVR} & \multirow{6}{*}{\shortstack{Add.(b) \\ Mult.(a) \\ MATH500(b) \\ ProofW.(b) \\ FOLIO(b)}} & \multirow{6}{*}{} & \multirow{6}{*}{} & \multirow{6}{*}{LOGIQA(a)} & \multirow{6}{*}{\textbf{83\%}}  \\  & & & & & & \\  & & & & & & \\ & & & & & & \\ & & & & & & \\ & & & & & & \\
\midrule
& & Avg. Ratio & 63\% & 3\% & 30\% & 3\% & 63\% \\
\bottomrule
\end{tabular}
\caption{Comparison of SCM type distributions between LLMs and LRMs. We assess the SCM types of different datasets across models and calculate the distributions for LLMs and LRMs, respectively. The first three columns present model type, model name, and primary training method. In columns 4-7, we report the SCM types of the tested mathematical and logical datasets, where SCM-I represents the ideal case. In the final column, we report the proportion of datasets classified as SCM-I for each model, with higher values indicating stronger causal reasoning capabilities. Additionally, rows 6 and 12 present the average proportions of different SCM types for LLMs and LRMs respectively, enabling fine-grained analysis of reasoning behaviors in these two model categories.}
\label{tab:main_scm}
\end{table*}

\section{Experiments}
We conduct extensive experiments with commercial and open-source LLMs and LRMs, testing the causal structures of the reasoning processes on different math and logical reasoning tasks.
\subsection{Experimental Settings}
\label{main_experimental_setting}
\subsubsection{Models} 
We test both non-reasoning models and reasoning models. As for non-reasoning models, we choose ChatGPT~\cite{openai2022chatgpt}, GPT-4~\cite{openai2023gpt4}, Llama2-7B/70B-Chat~\cite{touvron2023llama2} and Qwen-2.5-32B-Instruct~\cite{qwen2.5}. As for reasoning models, we select Deepseek-R1-0528~\cite{guo2025deepseek}, QwQ-32B~\cite{qwq32b}, DeepSeek-R1-Distill-Qwen-32/7B~\cite{guo2025deepseek} and DeepSeek-R1-Distill-Llama-8B~\cite{guo2025deepseek}. For Deepseek-R1, GPT3.5 and GPT4, we utilize the official API to conduct experiments; for other models, we deploy them using Vllm~\cite{kwon2023efficient}.

% Regarding model relationships, Qwen-2.5-32B-instruct builds upon the Qwen2.5-32B foundation through conventional instruction fine-tuning. DeepSeek-R1-Distill-Qwen-32B is constructed from the same Qwen2.5-32B base via reasoning model data distillation. QwQ-32B also leverages the Qwen2.5-32B architecture but incorporates outcome-based reinforcement learning.

\subsubsection{Datasets}
Our evaluation is conducted on mathematical reasoning and logical reasoning datasets. All selected datasets feature unambiguous ground-truth answers and necessitate chain-of-thought reasoning processes to reach the correct answers. The mathematical reasoning datasets consist of 3-digit multiplication, 9-digit addition, GSM8k~\cite{cobbe2021training} and MATH500~\cite{lightman2023lets}. The former two datasets, which were created in our previous work~\cite{bao2025likely}, evaluate fundamental arithmetic computation abilities, while MATH500 and GSM8k assess mathematical problem-solving skills through word problems. Each dataset above contains 500 samples. The logical reasoning datasets include ProofWriter~\cite{tafjord2020proofwriter}, FOLIO~\cite{han2022folio}, and LOGIQA\cite{liu2023logiqa}. ProofWriter and FOLIO are both multiple-choice datasets for evaluating deductive reasoning. We randomly draw 600 samples from the 5-hop reasoning development set of ProofWriter. As for FOLIO, we utilize all 204 instances from its development set. LOGIQA is a multiple-choice dataset designed to assess logical reasoning abilities in reading comprehension tasks. We extract 600 instances randomly from the LOGIQA2.0 test set.

\subsubsection{Inference Settings}
\label{main_inference_setting}
For experimental deployment, we conduct experiments with Deepseek-R1 via the official API, while other models are deployed using Vllm~\cite{kwon2023efficient}. We use a temperature setting of 0 for all experimental conditions unless specific parameters are recommended in the model's technical report. We set the maximum output length to 24,000 tokens, with any inference exceeding this limit being truncated. The details of the prompts can be found in our released code.

% \begin{figure}[t]
%     \centering
%     \subfloat[]{
%         \centering
%         \label{fig:cot0shot}
%         \includegraphics[width=0.45\textwidth]{Figures/cot0shot.png}
%     }
%     \vspace{-1em}
    
%     \subfloat[]{
%         \centering
%         \label{fig:newdirectbase}
%         \includegraphics[width=0.45\textwidth]{Figures/newdirectbase.png}
%     }
%     \vspace{-1em}
    
%     \subfloat[]{
%         \centering
%         \label{fig:newdirect}
%         \includegraphics[width=0.45\textwidth]{Figures/newdirect.png}
%     }
    
%     \caption{Examples of prompts using the Addition dataset. (a) Prompt for base models. Due to the unstable output of base models, complex instructions are employed for control. (b) Prompt for instruction-tuned models. Since instruction-tuned models do not automatically generate chain-of-thought (CoT) reasoning, "Please reason step by step" is incorporated to elicit the reasoning process. (c) Prompt for distilled and RL models. As these models inherently produce CoT reasoning, the simplest prompt is utilized.}
%     \label{fig:prompt}
%     \vspace{-2em}
% \end{figure}

\subsection{Deriving SCM from Treatment Experiments}
SCMs comprise nodes and edges. Nodes represent the variables in our framework, specifically $Z$, $X$, $Y$, and $T$, where $T$ appears exclusively in the LRM setting. Edges represent the causal relationships between variables, with $X$→$Y$ indicating that variable $X$ causally determines $Y$. Within each model category (LLM or LRM), the node set remains fixed, while different edge configurations define distinct SCM types. Edge presence is determined by intervention experiments, where a significant R-ATE indicates the existence of an edge between two nodes. After conducting treatment experiments on all edges under consideration to determine their existence, we obtain the SCM of the model on a specific task. As Table~\ref{tab:main_table} shows, GPT4 represents the 3-variable LLM with variables $Z$, $X$, and $Y$, while Deepseek-R1 represents the 4-variable LLM with variables $Z$, $T$, $X$, and $Y$.

We take Deepseek-R1 on the Mult. dataset as an example for showing how the SCMs are obtained. 
For the DeepSeek-R1 default condition, the model's accuracy on Mult. is 0.99, serving as the pre-intervention baseline. In the random Thinking, random CoT and Instruction bias conditions, we calculate the post-intervention accuracies following interventions on the Thinking, CoT, and Instruction variables, obtaining 0.988, 0.652, and 0.982, respectively. Using Equation~\ref{eq:rate}, we compute the R-ATE values for interventions on these three variables, obtaining 0.2\%, 34.1\%, and 0.8\%, respectively. According to McNemar's test~\cite{mcnemar1947note}, only the CoT intervention produces a statistically significant change, while interventions on Thinking and Instruction ($Z$) are not significant. Consequently, the SCM includes an edge from CoT ($X$) to Answer ($Y$), but excludes edges from Thinking ($T$) to Answer ($Y$) and Instruction ($Z$) to Answer. As shown in Figure~\ref{fig:SCM}, this SCM corresponds to Type-I SCM category (a).

% Specifically, DeepSeek-R1 on the Mult. dataset has only $X$ significantly affecting $Y$, with $T$ and $Z$ showing no significant effects to $Y$. This creates only one edge between $X$ and $Y$, corresponding to the Type I (a) SCM in Figure~\ref{fig:SCM}.

\subsection{Results of Causal Structures}
\label{sec:lrm_scms}
We separately calculate the SCM distributions for LLMs and LRMs in Table~\ref{tab:main_scm}. The findings are as follows. First, LLM models belonging to the ideal SCM-I average only 30\%, while the corresponding value for LRMs is 63\%. Notably, even the weakest model among LRMs possesses 3 type-I SCMs, which equals the maximum number of type-I SCMs found in any LLM model. This demonstrates that LRM models exhibit significantly stronger causality than LLMs.

Among LLMs, SCM-I accounts for the largest proportion, demonstrating a certain degree of causality. However, the SCM types are also relatively evenly distributed. Specifically, 20\% belong to SCM-II explaining and 27\% belong to SCM-IV memorizing. In both of these SCM types (combined 47\%), CoT cannot significantly influence the results, indicating that LLM's CoT is largely unfaithful. The remaining 23\% fall under SCM-III, which represents a mixture of reasoning and common case. SCM-II and SCM-III (combined 43\%), results are affected by ineffective conditions in the instruction, reflecting the inherent nature of autoregressive models.

For LRMs, the majority (63\%) of SCMs are ideal SCM-I, with the remaining most frequent type being SCM-III (30\%). This indicates that LRM's CoT has a very stable and significant influence on answers (combined 93\%). The remaining issues are concentrated on the influence of instruction-irrelevant conditions on results in SCM-III. Additionally, 3\% each belong to SCM-II and SCM-IV (combined 6\%), reflecting unfaithful CoT in a small number of cases, but this issue is significantly less prevalent than in LLMs. Notably, the vast majority of non-ideal cases among the tested LRMs occur in distilled LRMs, whereas QwQ-32B which relies almost entirely on RLVR, and DeepSeek-R1 which combines distillation with RLVR, exhibit almost no non-ideal SCMs, further demonstrating the positive impact of RLVR on causality. Moreover, a direct comparison among QwQ-32B (RLVR with thinking), R1-DeepSeek-Qwen-32B (SFT with thinking), and Qwen2.5-32B-Instruct (SFT without thinking) suggests that the presence or absence of Thinking ($T$) has limited effect on causality, whereas the training paradigm, particularly RLVR, plays the dominant role. Inspired by these observations, we further investigate the impact of different learning paradigms in Section~\ref{learning_tech}.

We further examine the four-variable SCM minor types for LRMs, where the distinction from three-variable SCMs lies in the decomposition of LRM's reasoning process into two separate variables: $T$ and $X$. Notably, the thinking variable significantly influences the answer in Type-I (b) and Type-III (a) and (c) configurations. The proportion of datasets belonging to these SCM categories is 67\% for both Qwen-based distilled models and 67\% for QwQ-32B, while R1-Distill-Llama-8B and DeepSeek-R1 show lower proportions at 33\% each. This indicates that the latter two LRMs exhibit more linear dependence on the preceding variable $X$ for answer generation, whereas the former three LRMs demonstrate simultaneous dependence on both $T$ and $X$ variables.

\subsection{Summary of Findings}
% Our experiments reveal that, although both distilled LRMs and RL-trained LRMs achieve substantial gains on accuracy-based metrics, their underlying causal structures diverge markedly. Across datasets, RL-trained LRMs consistently exhibit superior and more coherent SCMs, whereas distilled LRMs remain broadly similar to SFT models in their causal organization. These results indicate that distillation of long-chain CoT, by themselves, do not enhance the causal reasoning capabilities of LRMs. In contrast, RL plays a critical role in strengthening causal reasoning.
The CoT reasoning for LLMs is not causal but rather statistical. It is not only susceptible to interference from extraneous information in instructions, but also exhibit a substantial number of SCMs where CoT is unreliable. Our findings explain experimental observations from previous work showing that incorrect CoT can lead to correct conclusions, and vice versa~\cite{wu2024cofca, turpin2023language, paul2024making}. LRM models exhibit significantly superior causality compared to LLMs. The credibility of LRM's CoT is substantially higher than that of LLMs, though it is still somewhat influenced by the inherent autoregressive nature, causing answers to be disturbed by irrelevant information in instructions. Among LRMs, those trained with RLVR significantly outperform distilled LRMs, demonstrating the positive impact of RLVR on causality.

\begin{table*}[htbp]
\centering
\begin{tabular}{l|c|ccccccc|c|c}
\toprule
ICL & Metric & Add. & Mult. & GSM. & Pro. & FOL. & LQA. & AVG & SCM-I & Task Acc \\
\midrule
\multirow{3}{*}{0-shot} & R-ATE: X$\rightarrow$Y $\uparrow$ & $2.3\%_{F}$ & $100.0\%_{T}$ & $97.6\%_{T}$ & $17.9\%_{T}$ & $15.8\%_{T}$ & $8.3\%_{T}$ & $40.3\%$ & \multirow{3}{*}{2} & \multirow{3}{*}{0.572} \\
& R-ATE: Z$\rightarrow$Y $\downarrow$ & $70.8\%_{T}$ & $9.3\%_{T}$ & $0_{F}$ & $3.3\%_{T}$ & $4.3\%_{F}$ & $9.8\%_{T}$ & $16.2\%$ & & \\
& SCM & II & III & I & III & I & III & - & & \\
\midrule
\multirow{3}{*}{2-shot} & R-ATE: X$\rightarrow$Y $\uparrow$ & $0_{F}$ & $100.0\%_{T}$ & $99.2\%_{T}$ & $45.6\%_{T}$ & $46.6\%_{T}$ & $8.4\%_{T}$ & $\textbf{50.0\%}$ & \multirow{3}{*}{\textbf{3}} & \multirow{3}{*}{\textbf{0.598}} \\
& R-ATE: Z$\rightarrow$Y $\downarrow$ & $17.6\%_{T}$ & $0_{F}$ & $0_{F}$ & $19.9\%_{T}$ & $15.5\%_{T}$ & $1.9\%_{F}$ & 9.2\% &  & \\
& SCM & II & I & I & III & III & I & - & & \\
\midrule
\multirow{2}{*}{4-shot} & R-ATE: X$\rightarrow$Y $\uparrow$ & $0\%_{F}$ & $100.0\%_{T}$ & $99.5\%_{T}$ & $40.9\%_{T}$ & $39.0\%_{T}$ & $4.4\%_{F}$ & 47.3\% & \multirow{3}{*}{2}  & \multirow{3}{*}{0.580} \\
& R-ATE: Z$\rightarrow$Y $\downarrow$ & $0\%_{F}$ & $0\%_{F}$ & $0\%_{F}$ & $13.2\%_{T}$ & $20.4\%_{T}$ & $0\%_{F}$ & \textbf{5.7\%} & & \\
& SCM & IV & I & I & III & III & IV & - & & \\
\midrule
\multirow{2}{*}{8-shot} & R-ATE: X$\rightarrow$Y $\uparrow$ & $0\%_{F}$ & $99.7\%_{T}$ & $99.2\%_{T}$ & $47.1\%_{T}$ & $30.9\%_{T}$ & $6.8\%_{T}$ & 47.3\% & \multirow{3}{*}{\textbf{3}} & \multirow{3}{*}{0.592} \\
& R-ATE: Z$\rightarrow$Y $\downarrow$ & $0\%_{F}$ & $0\%_{F}$ & $0\%_{F}$ & $32.9\%_{T}$ & $25.2\%_{T}$ & $1.6\%_{F}$ & 10.0\% & & \\
& SCM & IV & I & I & III & III & I & - & & \\
\bottomrule
\end{tabular}
\caption{\emph{The impact of ICL} on causal relationships tested on \emph{GPT-3.5-Turbo}, where the best $|\text{R-ATE}|$ and task accuracy are marked in bold. The `{\it T/F}' indicates the statistical significance of the causal relation.}
\label{tab:icl-results}
\end{table*}

\section{Learning Techniques Shape Causal Structures}
\label{learning_tech}
Building on the previous section, we investigate how different learning paradigms shape the causal structures in language models. We systematically evaluate in-context learning (ICL), supervised fine-tuning (SFT), distillation, reinforcement learning (RL), and their combinations (SFT+RL and Distill+RL), assessing both causal structures and robustness. This allows us to quantify the contribution of each paradigm to causal alignment in reasoning models.

\subsection{Impact of In-Context Learning}
In-context learning (ICL) is widely adopted to elicit desired behaviors in LLMs, and is commonly used to trigger chain-of-thought reasoning to mimic human step-by-step inference. However, ICL techniques such as few-shot prompting are generally not recommended for LRMs~\cite{guo2025deepseek}, therefore we only evaluate the effectiveness of ICL on LLMs in our experiments.

\textit{Settings.} 
We evaluate ICL using GPT-3.5-Turbo as the base model. 
Following the same set of tasks described in Section~\ref{main_experimental_setting}, we prepend $k$ randomly sampled demonstrations to the prompt, with $k$ varied across conditions. All demonstrations are drawn from the development set without leakage into evaluation examples. For each condition, we conduct intervention experiments and compute SCM structures, R-ATEs, and task accuracies.

\textit{Results and Analysis.} As shown in Table~\ref{tab:icl-results}, compared to the zero-shot setting, ICL improves both task accuracy and causal alignment slightly. Specifically, ICL reduces the absolute R-ATE of the \textit{Instruction $\rightarrow$ Answer} edge while enhancing that of \textit{CoT $\rightarrow$ Answer}, leading to more coherent causal structures. This suggests that demonstrations help suppress spurious correlations and strengthen reasoning grounded in the CoT. 

\begin{table*}[htbp]
\centering
\begin{tabular}{l|c|ccccccc|c|c}
\toprule
Model & Metric & Add. & Mult. & GSM. & Pro. & FOL. & LQA. & AVG & SCM-I & Task Acc \\
\midrule
\multirow{3}{*}{Base} & R-ATE: X$\rightarrow$Y $\uparrow$ & $0\%_{F}$ & $100.0\%_{T}$ & $98.2\%_{T}$ & $32.0\%_{T}$ & $0\%_{F}$ & $9.0\%_{T}$ & $\textbf{39.9\%}$ & \multirow{3}{*}{\textbf{3}} & \multirow{3}{*}{\textbf{0.313}} \\
& R-ATE: Z$\rightarrow$Y $\downarrow$ & $100\%_{F}$ & $6.8\%_{F}$ & $2.3\%_{F}$ & $87.8\%_{T}$ & $24.4\%_{F}$ & $0\%_{F}$ & $36.9\%$ & & \\
& SCM & IV & I & I & III & IV & I & - & & \\
\midrule
\multirow{3}{*}{SFT} & R-ATE: X$\rightarrow$Y $\uparrow$ & $0_{F}$ & $92.3\%_{T}$ & $97.1\%_{T}$ & $0\%_{F}$ & $0\%_{F}$ & $5\%_{F}$ & $32.4\%$ & \multirow{3}{*}{0} & \multirow{3}{*}{0.300} \\
& R-ATE: Z$\rightarrow$Y $\downarrow$ & $100\%_{T}$ & $57.7_{T}$ & $9.5_{T}$ & $12.0\%_{T}$ & $34.6\%_{T}$ & $22.8\%_{T}$ & 39.4\% &  & \\
& SCM & II & III & III & II & II & II & - & & \\
\midrule
\multirow{2}{*}{DPO} & R-ATE: X$\rightarrow$Y $\uparrow$ & $0\%_{F}$ & $60.0\%_{F}$ & $96.3\%_{T}$ & $0\%_{F}$ & $12.3\%_{F}$ & $12.0\%_{T}$ & 30.1\% & \multirow{3}{*}{1}  & \multirow{3}{*}{0.275} \\
& R-ATE: Z$\rightarrow$Y $\downarrow$ & $0\%_{F}$ & $60.0\%_{F}$ & $8.6\%_{F}$ & $11.3\%_{F}$ & $8.6\%_{F}$ & $29.9\%_{T}$ & \textbf{19.7\%} & & \\
& SCM & IV & IV & I & IV & IV & III & - & & \\
\bottomrule
\end{tabular}
\caption{\emph{The impact of SFT/RLHF} on causal relationships based on \emph{Mistral-7B}, where SFT primarily enhances the causal connection between the instruction and answer, and DPO diminishes this relationship.}
\label{tab:sft-rlhf-results}
\end{table*}

\subsection{Impact of SFT and RLHF}
Supervised fine-tuning (SFT) enables LLMs to better follow human instructions, while reinforcement learning from human feedback (RLHF) further aligns model behavior with human preferences. However, recent studies have also shown that both SFT and RLHF may increase hallucinations and unfaithfulness~\cite{schulman2023reinforcement,yang2023alignment}. We therefore hypothesize that these paradigms may influence the underlying causal structures of reasoning.

\textit{Settings.} 
To validate this hypothesis, we analyze three models from the Mistral family: Mistral-7B-Base, Mistral-7B-SFT, and Mistral-7B-DPO~\cite{jiang2023mistral}. Since the base model cannot reliably follow instructions, we elicit its question-answering behavior using ICL with four demonstrations. For consistency, the same demonstrations are used for the SFT and DPO models. We then perform intervention experiments and compute SCM structures, R-ATEs, and task accuracies, following the same procedure as in Section~\ref{main_experimental_setting}.

\textit{Results and Analysis.}
As shown in Table~\ref{tab:sft-rlhf-results}, SFT generally weakens causal alignment, though individual models may exhibit specific variations. 
It reduces the average R-ATE on the \textit{CoT $\rightarrow$ Answer} edge while increasing that on the \textit{Instruction $\rightarrow$ Answer} edge, suggesting that SFT introduces spurious dependencies between instructions and answers, which in turn may cause hallucinations. 
In contrast, RLHF (via DPO) mitigates spurious correlations by reducing the strength of the \textit{Instruction $\rightarrow$ Answer} edge and lowering the average R-ATE from 36.9\% to 19.7\%, consistent with the human preference to separate answers from irrelevant instructions~\cite{ouyang2022training}. 
However, DPO also weakens the \textit{CoT $\rightarrow$ Answer} link (average R-ATE reduced to 0.138), indicating a trade-off where RLHF suppresses spurious patterns but may also diminish genuine causal connections.

\subsection{Impact of Distillation}
Distillation has become a dominant paradigm in recent LRMs, where smaller models learn the reasoning traces from stronger teachers. This approach is proved to yield stronger performance than RLVR training~\cite{guo2025deepseek}. Here, we analyze whether such gains also improve causal reasoning. Since distillation is technically a form of SFT, we compare it against instruction-tuning, another commonly used SFT approach.
% Distillation has become a dominant paradigm in recent LRMs, where smaller models learn the reasoning traces from stronger teachers. This approach is widely believed to yield stronger performance than directly applying RLVR to small models~\cite{guo2025deepseek}. Here, we analyze whether such gains also improve causal reasoning. 

\begin{table*}[htbp]
\centering
\begin{tabular}{l|c|ccccccc|c|c}
\toprule
Model & Metric & Add. & Mult. & MATH500 & Pro. & FOL. & LQA. & AVG & SCM-I & Task Acc \\
\midrule
\multirow{3}{*}{Qwen2.5-32B-Instr.} & R-ATE: X$\rightarrow$Y $\uparrow$ & $96.8\%_{T}$ & $84.5\%_{T}$ & $70.5\%_{T}$ & $44.4\%_{T}$ & $32.4\%_{T}$ & $6.7\%_{T}$ & $\textbf{55.9}\%$ & \multirow{3}{*}{3} & \multirow{3}{*}{0.76} \\
& R-ATE: Z$\rightarrow$Y $\downarrow$ & $19.7\%_{T}$ & $0_{F}$ & $8.6\%_{T}$ & $0_{F}$ & $1.4\%_{F}$ & $2.2\%_{T}$ & $5.3\%$ & & \\
& SCM & III & I & III & I & I & III & - & & \\
\midrule
\multirow{3}{*}{R1-Distill-Qwen-32B} & R-ATE: X$\rightarrow$Y $\uparrow$ & $95.6_{T}$ & $96.2\%_{T}$ & $97.1\%_{T}$ & $50.2\%_{T}$ & $18.6\%_{T}$ & $52.8\%_{T}$ & $53.3\%$ & \multirow{3}{*}{3} & \multirow{3}{*}{\textbf{0.84}} \\
& R-ATE: Z$\rightarrow$Y $\downarrow$ & $2.2\%_{F}$ & $2.3\%_{T}$ & $1.4_{F}$ & $6.1\%_{T}$ & $0.6\%_{F}$ & $2.6\%_{T}$ & \textbf{2.5}\% &  & \\
& SCM & I & III & I & III & I & III & - & & \\
\bottomrule
\end{tabular}
\caption{\emph{Comparison of Distill and SFT} on causal relationships based on \emph{Qwen-32B-Instruct} and \emph{R1-Distill-Qwen-32B}, where compared to Distill and Instruction-tuning, there are no significant causal changes.}
\label{tab:distill-results}
\end{table*}

\begin{table*}[htbp]
\centering
\begin{tabular}{|p{0.45\textwidth}|p{0.45\textwidth}|}
\hline
\textbf{Original Question} & \textbf{Modified Question} \\
\hline
If $f(x) = \frac{3x - 2}{x - 2}$, what is the value of $f(-2) + f(-1) + f(0)$? Express your answer as a common fraction. & 
If $f(x) = \frac{3x - 2}{x - 2}$, \textbf{and given that} $\boldsymbol{a = 7}$ \textbf{and} $\boldsymbol{b = -5}$, what is the value of $f(-2) + f(-1) + f(0)$? Express your answer as a common fraction. \\
\hline
\end{tabular}
\caption{Comparison of Original and Modified Questions. The bold portion ``and given that a=7 and b=-5" represents extraneous conditions irrelevant to problem solving, added in MATH500-NOOP to test whether models rely on superficial features.}
\label{tab:math500_noop_example}
\end{table*}

\begin{table*}[t]
\centering
\begin{tabular}{|c|c|c|}
\hline
 & \textbf{Math500 Acc: High} & \textbf{Math500 Acc: Low} \\
\hline
\textbf{Math500-Noop Acc: High} & 
\begin{tabular}{@{}c@{}}
Good fitting, few spurious features, \\
strong generalization, \\
\textbf{strong causality}
\end{tabular} & 
\begin{tabular}{@{}c@{}}
Rare state, \\
not discussed
\end{tabular} \\
\hline
\textbf{Math500-Noop Acc: Low} & 
\begin{tabular}{@{}c@{}}
Good fitting, many spurious features, \\
poor generalization, \\
\textbf{weak causality}
\end{tabular} & 
\begin{tabular}{@{}c@{}}
Underfitting, few spurious features \\
but also few genuine features, \\
\textbf{weak causality}
\end{tabular} \\
\hline
\end{tabular}
\caption{Analysis of Model Performance Across Different Feature Combinations}
\label{tab:feature_analysis}
\end{table*}

\begin{table*}[htbp]
\centering\scriptsize
\renewcommand{\arraystretch}{1.5}
\begin{tabular}{>{\centering\arraybackslash}m{1.3cm}>{\centering\arraybackslash}m{2.0cm}>{\centering\arraybackslash}m{5.5cm}>{\centering\arraybackslash}m{1.2cm}>{\centering\arraybackslash}m{1.6cm}>{\centering\arraybackslash}m{0.8cm}>{\centering\arraybackslash}m{2.8cm}} \\
\toprule
\textbf{Group} & \textbf{Base Model} & \textbf{Comparison Setting} & \textbf{Source} & \textbf{Algorithm} & \textbf{Dataset} & \textbf{Purpose} \\
\midrule
Base-RLVR & Qwen2.5-3B-Base & \begin{tabular}{@{}c@{}}
Qwen2.5-3B-Base \\
Qwen2.5-3B-Base-GRPO-600/1000/2000 steps
\end{tabular} & \begin{tabular}{@{}c@{}}
Open source \\
Self-trained
\end{tabular} & \begin{tabular}{@{}c@{}}- \\GRPO\end{tabular} & \begin{tabular}{@{}c@{}}- \\DeepScaleR\end{tabular} & Base vs Base+RLVR \\
\midrule
SFT-RLVR & Qwen2.5-3B-Instruct & \begin{tabular}{@{}c@{}}
Qwen2.5-3B-Instruct \\
Qwen2.5-3B-Instruct-GRPO-200/1000/2000 steps
\end{tabular} & \begin{tabular}{@{}c@{}}
Open source \\
Self-trained
\end{tabular} & \begin{tabular}{@{}c@{}}
Instr. \\ GRPO\end{tabular} & \begin{tabular}{@{}c@{}}
- \\ DeepScaleR\end{tabular} & SFT vs SFT+RL \\
\midrule
Base-RLVR/Distill & Qwen2.5-3B-Base & \begin{tabular}{@{}c@{}}
Qwen2.5-3B-Base-Distill-Openr1 \\
Qwen2.5-3B-Base-GRPO-Openr1 
\end{tabular} & \begin{tabular}{@{}c@{}}
Self-trained \\
Self-trained
\end{tabular} & \begin{tabular}{@{}c@{}}
Distill \\
GRPO
\end{tabular} & \begin{tabular}{@{}c@{}}
OpenR1 \\
OpenR1
\end{tabular} & Base+RL vs Base+Distill \\
\midrule
Distill-RLVR & Qwen2.5-Math-1.5B & \begin{tabular}{@{}c@{}}
Deepseek-Distill-Qwen-1.5B \\
DeepscaleR-1.5B-Preview 
\end{tabular} & \begin{tabular}{@{}c@{}}
Open source \\
Open source
\end{tabular} & \begin{tabular}{@{}c@{}}
Distill \\
Distill+GRPO
\end{tabular} & \begin{tabular}{@{}c@{}}
- \\
-
\end{tabular} & Distill vs Distill+RL \\
\bottomrule
\end{tabular}
\caption{We utilize four model groups to elucidate why RLVR improves causal reasoning in models. Details regarding the source, algorithm, dataset, and experimental objectives for each group are provided in this table.}
\label{tab:analysis setting}
\end{table*}

\textit{Settings.}
We analyze three LRMs distilled with DeepSeek-R1 data: \textit{R1-Distill-Qwen-7B}, \textit{R1-Distill-Qwen-32B}, and \textit{R1-Distill-Llama-8B}, and compare them against the Instruction-tuned model \textit{Qwen2.5-32B-Instruct}. Notably, both \textit{R1-Distill-Qwen-32B} and \textit{Qwen2.5-32B-Instruct} are derived from the same base model Qwen2.5-32B.

\textit{Results and Analysis.}
% As shown in Table~\ref{tab:distill-results}, all distilled models achieve only 3/6 ideal (type-I) SCMs, whereas QwQ-32B achieves 6/6. This indicates that distillation mainly transfers surface-level reasoning and often amplifies spurious correlations, while RLVR better suppresses spurious features and strengthens causal alignment. In summary, distillation yields strong accuracy but limited causal reasoning, with RLVR showing clear advantages in aligning reasoning with ideal causal structures.
As shown in Table~\ref{tab:main_table}, the three distilled LRMs exhibit same proportions of Ideal SCMs, all achieving 50\%, indicating consistent performance across distilled LRMs. We further compare Qwen2.5-32B-Instruct and R1-Distill-Qwen-32B, with detailed results presented in Table~\ref{tab:distill-results}. The results show that distillation significantly improves average task accuracy compared to instruction-tuning, which aligns with findings from other studies~\cite{guo2025deepseek}. However, both models have an identical number (3) of type-I SCMs, and the R-ATE changes for X→Y and Z→Y edges are minimal. This suggests that while distilled LRMs enhance task performance, they do not improve causality, showing no distinction from conventional instruction-tuning at the causal level.

\subsection{Impact of RLVR}
Reinforcement learning from verifiable rewards (RLVR) has recently emerged as a key optimization paradigm in LRMs, driving substantial gains in reasoning performance. Unlike RLHF, which relies on subjective human preference signals, RLVR leverages automatically verifiable outcomes, the correctness of the final answer, as reward signals.  By directly optimizing models with respect to correctness, RLVR not only improves task accuracy but also refines the reasoning process, making it more consistent with causal reasoning structures.  In this subsection, we analyze how RLVR shapes causality under the following SFT+GRPO settings.

\textit{Settings.}
We apply RLVR using the GRPO algorithm on Qwen2.5-3B-Instruction, following the outcome reward method of DeepScaleR~\cite{deepscaler2025}.
Given the relatively small scale of the model, we evaluate it on 6-digit Addition, 2-digit Multiplication, Math500, ProofWriter, FOLIO and LOGIQA. The DeepScaleR dataset consists of math word problems similar to MATH500, making MATH500 an in-domain task. Addition and Multiplication can be considered out-of-distribution (OOD) arithmetic problems, while ProofWriter, FOLIO and LOGIQA represent OOD logical reasoning problems.

\textit{Results and Analysis.}
% The results in Table~\ref{tab:grpo-results} show that RLVR progressively enhances causal reasoning in the base model, though with some instability due to exploration. The number of ideal (type-I) SCMs increases from none at initialization to 1 at step 600, dips to 0 at step 1000, and recovers to 2 at step 2000. Even at checkpoints with fewer ideal SCMs, R-ATE values are consistently higher than the base model, suggesting that RLVR strengthens causal connections before they reach statistical significance. 
According to Table~\ref{tab:grpo-results}, GRPO improves task accuracy from 0.59 to 0.62. Simultaneously, the number of Type-I SCMs increases from 1 to 4. Among all the methods discussed above, GRPO achieves the largest gain in causality. This causal enhancement can also be observed in Table~\ref{tab:main_scm}. QwQ-32B-Instruct achieves Type-I SCMs in 5 tasks, substantially outperforming R1-Distill-Qwen-32B and Qwen2.5-32B, which employ distillation or instruction-tuning on the same base model.

Beyond the in-domain MATH500, \textit{Qwen2.5-3B-Instruct-GRPO} demonstrates substantial out-of-distribution gains on logical reasoning benchmarks, despite being trained exclusively on mathematical data. As shown in Table~\ref{tab:grpo-results}, the baseline prior to GRPO achieves a Type-I SCM only on FOLIO, whereas GRPO upgrades \emph{all three} OOD logic tasks (ProofWriter, FOLIO, LOGIQA) to Type-I. This improvement is primarily driven by suppressing the spurious $Z\!\rightarrow\!Y$ edge—from $11.2\%$, $3.8\%$, and $3.1\%$ to $1.5\%$, $0\%$, and $1.9\%$ on Pro./FOL./LQA., respectively—while maintaining strong $X\!\rightarrow\!Y$ links across tasks. These results provide compelling evidence that RLVR learns transferable causal patterns rather than memorizing task-specific correlations~\cite{chu2025sft}.

% In addition to the in-domain task MATH500, Qwen2.5-3B-Instruction-GRPO generalizes effectively to all OOD logical reasoning benchmark, despite being trained only on mathematical data. This indicates that RLVR captures genuine causal patterns that extend beyond task-specific correlations.

% In summary, applying RLVR significantly improves causal reasoning on in-domain tasks and generalizes to other OOD tasks.
% In summary, applying RLVR improves causal alignment, enhances robustness to spurious features, and supports cross-domain generalization. These findings establish RLVR as a powerful method for strengthening reasoning causality in language models.

\begin{table*}[htbp]
\centering
\begin{tabular}{@{}l|c|ccccccc|c|c@{}}
\toprule
Model & Metric & Add. & Mult. & MATH500 & Pro. & FOL. & LQA. & AVG & SCM-I & Task Acc \\
\midrule
\multirow{3}{*}{Qwen2.5-3B-Instr.} & R-ATE: X$\rightarrow$Y $\uparrow$ & $69.3\%_{T}$ & $72.3\%_{T}$ & $90.2\%_{T}$ & $49.9\%_{T}$ & $58.1\%_{T}$ & $27.3\%_{T}$ & $\textbf{61.2}\%$ & \multirow{3}{*}{1} & \multirow{3}{*}{0.59} \\
& R-ATE: Z$\rightarrow$Y $\downarrow$ & $7.2\%_{T}$ & $6.3\%_{T}$ & $13.1\%_{T}$ & $11.2\%_{T}$ & $3.8\%_{F}$ & $3.1\%_{T}$ & $7.5\%$ & & \\
& SCM & III & III & III & III & I & III & - & & \\
\midrule
\multirow{3}{*}{Qwen2.5-3B-Instr-GRPO} & R-ATE: X$\rightarrow$Y $\uparrow$ & $93.8_{T}$ & $72.4\%_{T}$ & $72.9\%_{T}$ & $39.2\%_{T}$ & $46.0\%_{T}$ & $27.7\%_{T}$ & $58.7\%$ & \multirow{3}{*}{\textbf{4}} & \multirow{3}{*}{\textbf{0.62}} \\
& R-ATE: Z$\rightarrow$Y $\downarrow$ & $20.4\%_{T}$ & $9.0\%_{T}$ & $0_{F}$ & $1.5\%_{F}$ & $0\%_{F}$ & $1.9\%_{F}$ & \textbf{5.5}\% &  & \\
& SCM & III & III & I & I & I & I & - & & \\
\bottomrule
\end{tabular}
\caption{\emph{The impact of RLVR} on causal relationships based on \emph{Qwen2.5-3B-Instruct}, where GRPO significantly increased the number of good SCMs and improved task performance.}
\label{tab:grpo-results}
\end{table*}

\section{How Does RLVR Enhance Causality}
\label{sec:rlvr_process}
The analyses in the previous section demonstrate that RLVR enhances causal reasoning structures significantly. Building on these observations, we now aim to explain how RLVR leads to ideal causality. Intuitively, causality is closely tied to the contrast between genuine and spurious features: genuine features are causally related to the true label, while spurious features lack causal connection to the label but are coincidentally correlated with it in the data~\cite{zhou2024towards, wu2024causality}. Models exhibit better causal alignment when reasoning relies more on genuine features and less on spurious ones~\cite{zhou2024towards}. To validate this hypothesis, we design three sets of comparative experiments.

\begin{figure*}[t]
    \centering
    \subfloat[]{\includegraphics[width=0.49\linewidth]{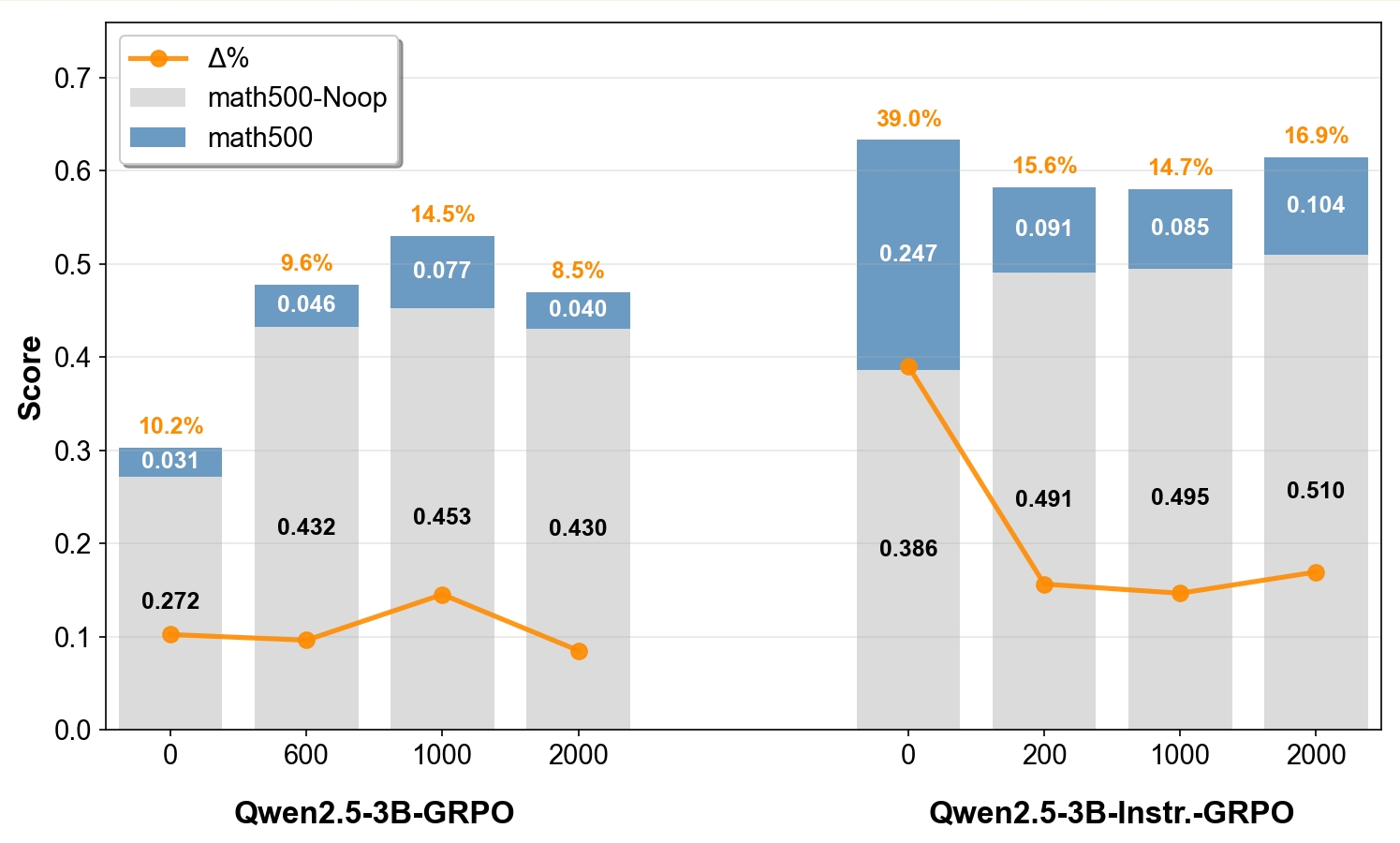}
    \label{fig:noop_process}}
    \hfill
    \subfloat[]{\includegraphics[width=0.49\linewidth]{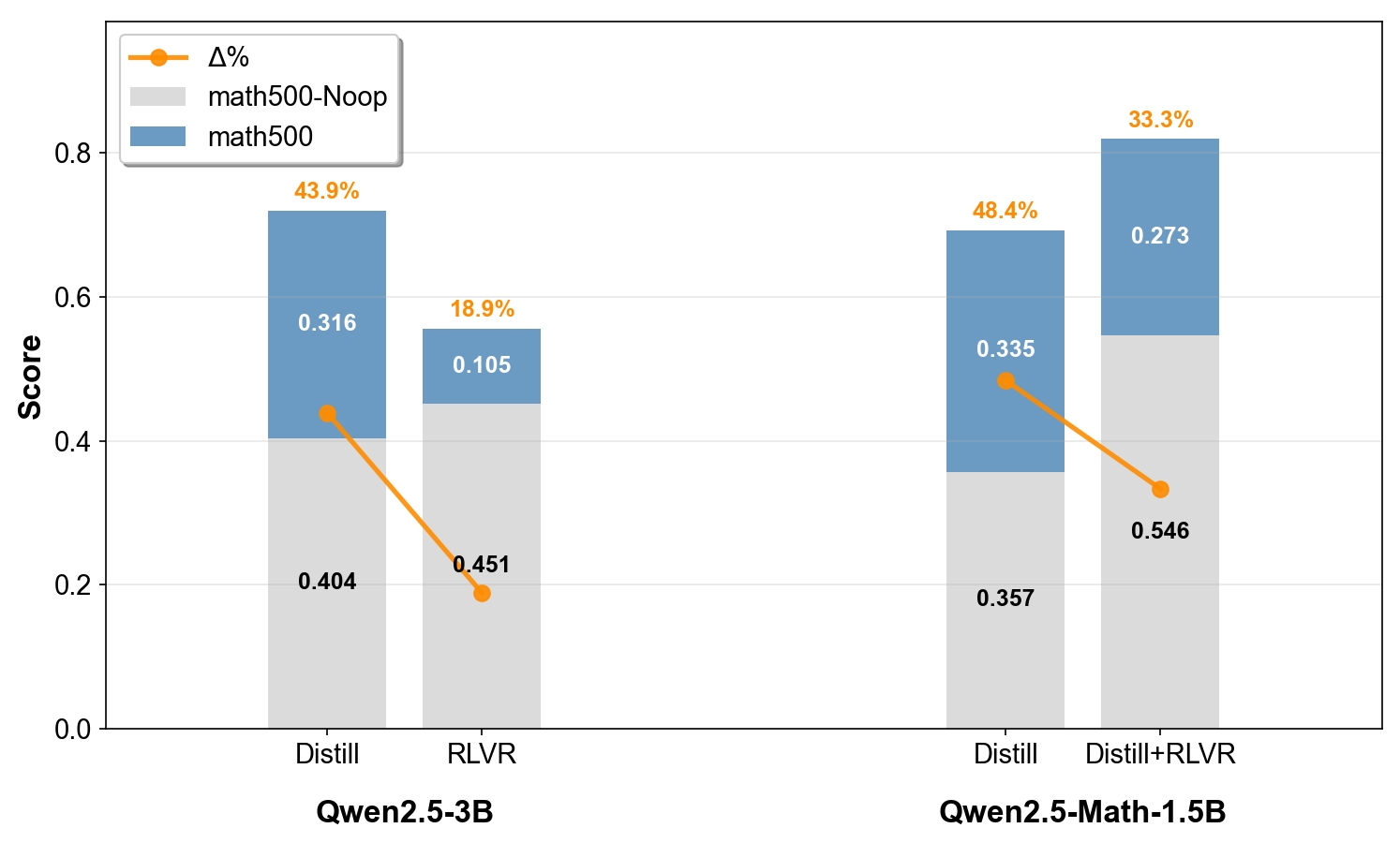}
    \label{fig:noop_distill}}
    \caption{We evaluate the performance of four groups of models: Base-RL, SFT-RL, Base-RL/Distill, and Distill-RL, on Math500 and Math500-Noop. In the Base-RLVR group, 0, 600, 1000, and 2000 represent Qwen2.5-3B and checkpoints of Qwen2.5-3B after 600, 1000, and 2000 GRPO steps, respectively. In the SFT-RLVR group, 0, 200, 1000, and 2000 represent Qwen2.5-3B-Instruct and checkpoints of Qwen2.5-3B-Instruct after 200, 1000, and 2000 GRPO steps, respectively. In the Base-RLVR/Distill group, ``Distill'' represents the Qwen2.5-3B model distilled with OpenR1-Math-220k data, while ``RLVR'' represents the model obtained through GRPO using the same base architecture. In the Distill-RLVR group, ``Distill'' refers to the open-source Deepseek-Distill-Qwen-1.5B model, which is trained from Qwen2.5-Math-1.5B using distillation, while ``Distill+RLVR'' represents the open-source DeepscaleR-1.5B-Preview model, which applies GRPO on top of Deepseek-Distill-Qwen-1.5B. The bar chart represents the accuracy of Math500 and Math500-Noop, while the line chart depicts the relative difference between Math500 and Math500-Noop. We utilize Math500-Noop to measure the model's generalization capability and the extent to which genuine features are learned; the relative gap in accuracy between Math500 and Math500-Noop is employed to quantify the degree of spurious feature learning. (a) Contains Base-RLVR and SFT-RLVR, which reflect the changes in genuine and spurious features when applying GRPO starting from Base and SFT models, respectively. (b) Contains Base-RLVR/Distill and Distill-RLVR. The former reflects the changes in genuine and spurious features between RLVR and Distill under identical training data, while the latter reflects the scenario of applying RLVR on top of the Distill model.}
    \label{fig:noop}
\end{figure*}

\begin{table*}[h]
\centering\scriptsize
\begin{tabular}{@{}l|l|c|ccc|cccc|c@{}}
\hline
\multirow{2}{*}{Model} & \multirow{2}{*}{Metric} & indomain & \multicolumn{3}{c|}{OOD-Arithmetic} & \multicolumn{4}{c|}{OOD-logic} & \multirow{2}{*}{Good SCM(I)} \\
\cline{3-10}
 &  & Math500 & Add & Mult & avg & ProofWriter & FOLIO & LOGIQA & avg &  \\
\hline
\multirow{4}{*}{Qwen2.5-3B-Base} & CoT$\rightarrow$Ans: R-ATE$\uparrow$ & $79.39\%_{T}$ & $5.06\%_{F}$ & $47.95\%_{T}$ & $26.51\%$ & $39.31\%_{T}$ & $63.06\%_{T}$ & $12.42\%_{T}$ & $38.26\%$ & - \\ 
 & Instr$\rightarrow$Ans: R-ATE$\downarrow$ & $17.94\%_{T}$ & $94.90\%_{T}$ & $87.67\%_{T}$ & $91.29\%$ & $46.82\%_{T}$ & $18.92\%_{T}$ & $69.57\%_{T}$ & $45.10\%$ & - \\
 & SCM & III & II & III & - & III & III & III & - & 0 \\
\hline
\multirow{4}{*}{(GRPO step 600)} & CoT$\rightarrow$Ans: R-ATE$\uparrow$ & $79.52\%_{T}$ & $98.20\%_{T}$ & $96.65\%_{T}$ & $97.43\%$ & $48.21\%_{T}$ & $62.04\%_{T}$ & $22.06\%_{T}$ & $44.10\%$ & - \\
 & Instr$\rightarrow$Ans: R-ATE$\downarrow$ & $5.85\%_{T}$ & $44.01\%_{T}$ & $10.05\%_{T}$ & $27.03\%$ & $6.92\%_{T}$ & $1.76\%_{F}$ & $22.06\%_{T}$ & $10.25\%$ & - \\
 & SCM & III & III & III & - & III & I & III & - & 1 \\
\hline
\multirow{4}{*}{(GRPO step 1000)} & CoT$\rightarrow$Ans: R-ATE$\uparrow$ & $74.51\%_{T}$ & $100.00\%_{T}$ & $84.87\%_{T}$ & $92.44\%$ & $39.07\%_{T}$ & $51.49\%_{T}$ & $27.00\%_{T}$ & $39.19\%$ & - \\
 & Instr$\rightarrow$Ans: R-ATE$\downarrow$ & $6.77\%_{T}$ & $17.84\%_{T}$ & $3.59\%_{T}$ & $10.72\%$ & $11.55\%_{T}$ & $10.69\%_{T}$ & $16.25\%_{T}$ & $12.83\%$ & - \\
 & SCM & III & III & III & - & III & III & III & - & 0 \\
\hline
\multirow{4}{*}{(GRPO Step 2000)} & CoT$\rightarrow$Ans: R-ATE$\uparrow$ & $87.73\%_{T}$ & $97.79\%_{T}$ & $80.36\%_{T}$ & $89.08\%$ & $58.32\%_{T}$ & $53.90\%_{T}$ & $11.34\%_{T}$ & $41.19\%$ & - \\
 & Instr$\rightarrow$Ans: R-ATE$\downarrow$ & $0.20\%_{F}$ & $59.38\%_{T}$ & $16.54\%_{T}$ & $37.96\%$ & $4.93\%_{T}$ & $0.89\%_{F}$ & $10.42\%_{T}$ & $5.41\%$ & - \\
 & SCM & I & III & III & - & III & I & III & - & 2 \\
\hline
\multirow{4}{*}{Qwen2.5-3B-Instr} & CoT$\rightarrow$Ans: R-ATE$\uparrow$ & $90.18\%_{T}$ & $69.28\%_{T}$ & $72.30\%_{T}$ & $70.79\%$ & $49.91\%_{T}$ & $58.07\%_{T}$ & $27.29\%_{T}$ & $45.09\%$ & - \\
 & Instr$\rightarrow$Ans: R-ATE$\downarrow$ & $13.09\%_{T}$ & $7.23\%_{T}$ & $6.33\%_{T}$ & $6.78\%$ & $11.19\%_{T}$ & $3.80\%_{F}$ & $3.05\%_{T}$ & $6.01\%$ & - \\
 & SCM & III & III & III & - & III & I & III & - & 1 \\
\hline
\multirow{4}{*}{(GRPO Step 200)} & CoT$\rightarrow$Ans: R-ATE$\uparrow$ & $69.57\%_{T}$ & $99.78\%_{T}$ & $78.90\%_{T}$ & $89.34\%$ & $15.23\%_{T}$ & $54.39\%_{T}$ & $0\%_{F}$ & $23.21\%$ & - \\
 & Instr$\rightarrow$Ans: R-ATE$\downarrow$ & $2.80\%_{F}$ & $9.89\%_{T}$ & $8.12\%_{T}$ & $9.01\%$ & $21.87\%_{T}$ & $2.49\%_{F}$ & $1.29\%_{F}$ & $8.55\%$ & - \\
 & SCM & I & III & III & - & III & I & IV & - & 2 \\
\hline
\multirow{4}{*}{(GRPO Step 1000)} & CoT$\rightarrow$Ans: R-ATE$\uparrow$ & $75.99\%_{T}$ & $94.59\%_{T}$ & $75.07\%_{T}$ & $84.83\%$ & $39.18\%_{T}$ & $58.54\%_{T}$ & $34.95\%_{T}$ & $44.22\%$ & - \\
 & Instr$\rightarrow$Ans: R-ATE$\downarrow$ & $0\%_{F}$ & $58.11\%_{T}$ & $8.80\%_{T}$ & $33.46\%$ & $2.04\%_{F}$ & $0.83\%_{F}$ & $6.81\%_{T}$ & $3.23\%$ & - \\
 & SCM & I & III & III & - & I & I & III & - & 3 \\
\hline
\multirow{4}{*}{(GRPO Step 2000)} & CoT$\rightarrow$Ans: R-ATE$\uparrow$ & $72.93\%_{T}$ & $93.75\%_{T}$ & $72.38\%_{T}$ & $83.07\%$ & $39.23\%_{T}$ & $46.04\%_{T}$ & $27.66\%_{T}$ & $37.64\%$ & - \\
 & Instr$\rightarrow$Ans: R-ATE$\downarrow$ & $0\%_{F}$ & $20.43\%_{T}$ & $9.01\%_{T}$ & $14.72\%$ & $1.47\%_{F}$ & $0\%_{F}$ & $1.87\%_{F}$ & $1.11\%$ & - \\
 & SCM & I & III & III & - & I & I & I & - & 4 \\
\hline
\end{tabular}
\caption{The number of type-I SCMs generally increases during RL training process.}
\label{tab:RL_process}
\end{table*}

\subsection{Measuring Spurious Feature Reliance}
To quantify the extent to which models exploit spurious correlations, we construct a variant of Math500 following the GSM-Noop methodology~\cite{gsm-symbolic}. The dataset, termed \textit{Math500-Noop}, augments each problem with 1–2 irrelevant numerical conditions generated by GPT-4.1-mini. An example is shown in Table~\ref{tab:math500_noop_example}. Since the added conditions are causally unrelated to the solution, a model that relies on pattern matching or shortcut heuristics may be misled, whereas a model that performs genuine reasoning should remain robust.
We then define the metric:
\begin{equation}
    \Delta\% = \frac{Acc(\text{Math500})-Acc(\text{Math500-Noop})}{Acc(\text{Math500})}
\end{equation}
with larger $\Delta\%$ indicating stronger spurious reliance.
Table \ref{tab:feature_analysis} summarizes how combinations of Math500 and Math500-Noop accuracies map to causality levels: strong causality requires both high fitting (high Acc on Math500 and Math500-Noop) and low spurious reliance (small $\Delta\%$), while other cases correspond to weak causality.

\subsection{Correlation Between Causality and Spurious Features}
\label{sec:rl_extent}
We now investigate whether the CoT causality and the proportion of the spurious features are quantitatively correlated during RLVR training. Under the assumption that stronger causality arises from reduced reliance on spurious features, metrics of causal alignment and spurious feature dependence should exhibit consistent correlation across training checkpoints. To validate this, we explicitly define how to measure spurious feature reliance and causal alignment, and then compute their correlation over multiple RLVR training trajectories.

\textit{Settings.}
We analyze two model groups: Base-RL, SFT-RL. For each group, we collect checkpoints along the RLVR training trajectory and compute two complementary metrics. The first is the number of type-I SCMs, derived from the intervention-based SCM framework in Section~\ref{main_experimental_setting}, which reflects the extent of causal alignment. The second is the relative accuracy difference $\Delta\%$ between Math500 and Math500-Noop, which serves as a proxy for spurious feature reliance. To capture changes induced by RLVR rather than absolute levels, we compute differences for both metrics relative to the initial (pre-RLVR) checkpoint of each trajectory. Finally, we pool all checkpoints from Base-RL and SFT-RL and calculate the Pearson correlation coefficient between these change values. The corresponding models and other details are provided in Table~\ref{tab:analysis setting}.

\textit{Results and Analysis.} 
Across both Base-RL and SFT-RL trajectories, we observe that causal alignment and spurious feature reliance evolve in opposite directions during RLVR training. For the Base-RL group, the number of ideal SCMs fluctuates across checkpoints (Table~\ref{tab:RL_process}), rising from 0 to 1 at step 600, dropping back to 0 at step 1000, and then recovering to 2 at step 2000. Interestingly, these fluctuations are mirrored by changes in spurious feature reliance ($\Delta\%$, Fig.~\ref{fig:noop}a): when SCM counts increase, $\Delta\%$ decreases, and vice versa. This pattern suggests that improvements in causal alignment are consistently accompanied by reductions in spurious correlations. The SFT-RL group shows a more stable trajectory: starting with 1 ideal SCM in the baseline, the count increases steadily to 4 after 2000 RLVR steps, while $\Delta\%$ decreases from 39\% to around 17\%, confirming that RLVR simultaneously strengthens causal structures and suppresses spurious features. 

To quantify this relationship, we compute the Pearson correlation between changes in type-I SCM counts and changes in $\Delta\%$, relative to the pre-RLVR baseline. 
Pooling checkpoints from both Base-RL and SFT-RL, we obtain a coefficient of $-0.68$($p=0.065$), indicating a strong negative correlation that is marginally significant. The consistent trend across trajectories provides robust evidence that improvements in causality under RLVR are tightly linked to the suppression of spurious features.

These findings reinforce our hypothesis that models achieve stronger causal alignment only when reasoning relies more on genuine features and less on superficial correlations. In this sense, Math500-Noop serves as a critical probe—its performance reflects the extent of genuine reasoning, while the gap to Math500 quantifies spurious reliance. Importantly, our results also suggest that neither abundant genuine features alone nor reduced spurious features alone are sufficient; effective causality emerges only when both conditions are simultaneously met.

\subsection{Distillation vs. RLVR}
Distillation has recently been advocated such as Deepseek-R1~\cite{guo2025deepseek} as a more effective strategy for training smaller reasoning models than RLVR.
From a causal perspective, however, it remains unclear whether the gains from distillation reflect genuine reasoning improvements or amplification of spurious correlations. To answer this, we compare distillation and RLVR under controlled conditions and further examine whether RLVR can mitigate the spurious features inherited from distilled checkpoints.

\textit{Settings.}
We consider two setups for comparing distillation and RLVR.
First, in a controlled comparison, we train \textit{Qwen2.5-3B} on OpenR1-Math-220k with both distillation and RLVR, obtaining \textit{Qwen2.5-3B-Base-Distill-OpenR1} and \textit{Qwen2.5-3B-Base-GRPO-OpenR1}. This setting enables a direct evaluation of how the two paradigms affect causal alignment under identical data conditions.
Second, to test whether RLVR can refine distilled models, we analyze two open-source checkpoints: \textit{Deepseek-Distill-Qwen-1.5B} and its RLVR-enhanced counterpart \textit{Deepscaler-1.5B-Preview}~\cite{deepscaler2025}. We present more details of these two groups of models in Table~\ref{tab:analysis setting}.

% Evaluation uses the same intervention-based SCM framework described in Section~\ref{main_experimental_setting} to measure causal alignment, where the number of type-I SCMs and edge-level R-ATEs are the key indicators.
To assess reliance on spurious features, we employ the \textit{Math500-Noop} dataset introduced in Section~\ref{sec:rl_extent}, and compute the relative accuracy gap between Math500 and Math500-Noop ($\Delta\%$) as a proxy for spurious feature dependence. This ensures methodological consistency with the analyses in the previous subsection. 

\textit{Results and Analysis.}
In the controlled setting (Base-Distill/RL), we find a clear divergence between the two paradigms. Distillation substantially improves in-distribution accuracy (e.g., Math500), but its performance on Math500-Noop lags behind RLVR despite the latter’s lower Math500 accuracy (Figure~\ref{fig:noop}b). This indicates that distillation’s gains are largely driven by spurious correlations.%, while RLVR captures more genuine reasoning features and yields stronger causal alignment, as evidenced by higher type-I SCM counts and more coherent R-ATEs (Table~\ref{tab:RL_process}).

In the Distill-RL group, comparing \textit{Deepseek-Distill-Qwen-1.5B} with \textit{Deepscaler-1.5B-Preview}, we observe that RLVR improves both Math500 and Math500-Noop performance, narrowing the gap between them (Figure~\ref{fig:noop}b). This demonstrates that RLVR can reduce some of the spurious correlations introduced by distillation and enhance causal alignment beyond the distilled baseline.

Overall, while distillation yields strong raw accuracy, it does so at the cost of amplifying spurious features. In contrast, RLVR enhances causal robustness by diminishing reliance on spurious correlations, whether applied directly to base models or used to refine distilled ones.

\subsection{Discussion}
Our analyses confirm that robust reasoning requires not only strong task performance but also reliable causal structures. Building on this, we highlight three main insights into the role of RLVR and distillation in shaping causal structures of reasoning models.  

First, RLVR improves generalization by systematically reducing reliance on spurious features. This is consistent with prior analyses suggesting that verifiable reward signals encourage models to explore diverse reasoning paths and suppress shortcut heuristics~\cite{deng2025trial, chen2025reasoningerasurveylong}. Importantly, such improvements are not fully reflected in standard task accuracy but become evident on causality-sensitive benchmarks such as Math500-Noop, highlighting the necessity of causal evaluation as a complement to conventional metrics.

Second, distilled models exhibit substantially higher levels of spurious features (Figure~\ref{fig:noop}b). Even after applying RLVR on top of distillation, spurious correlations persist abundantly, suggesting that correlations introduced during distillation are partially irreversible. This limitation indicates that relying solely on accuracy-based evaluation can be misleading: distillation appears effective on in-distribution tasks but weakens robustness in ways that persist even after further RLVR training.  

Third, our findings reveal a fundamental trade-off between task accuracy and causal robustness. Distillation consistently achieves higher in-distribution accuracy but amplifies spurious correlations, while RLVR sacrifices some accuracy yet produces stronger causal alignment and robustness under distribution shifts. This raises an open challenge for future work: to design training paradigms that simultaneously achieve strong fitting ability and causal reliability, for example by combining the strengths of distillation and RLVR in hybrid frameworks or by explicitly incorporating causal signals into optimization objectives.

\section{Conclusion}
We investigate the causal structure of CoT reasoning for LLMs and LRMs. Our findings reveal that LLMs exhibit statistical rather than causal reasoning, whereas LRMs demonstrate superior causality in their reasoning processes.
Our experimental evidence shows that RLVR training progressively strengthens genuine causal relationships while mitigating spurious correlations, establishing a clear pathway for developing more reliable reasoning systems. These findings provide crucial insights for causality-driven AI development and highlight reinforcement learning as the preferred training approach when causally robust reasoning is essential for trustworthy AI applications.

\bibliographystyle{IEEEtran}
\bibliography{bare_jrnl_new_sample4}

\vfill

\end{document}